\documentclass{article}

    \PassOptionsToPackage{numbers, compress}{natbib}

\usepackage[preprint]{neurips_2024}

\usepackage[utf8]{inputenc} 
\usepackage[T1]{fontenc}    
\usepackage{hyperref}       
\usepackage{url}            
\usepackage{booktabs}       
\usepackage{amsfonts}       
\usepackage{nicefrac}       
\usepackage{microtype}      
\usepackage{xcolor}         
\usepackage[letterpaper]{geometry}
\usepackage{wrapfig} 
\usepackage[american]{babel}

\usepackage[numbers]{natbib}
\usepackage{mathtools} 
\usepackage{booktabs} 
\usepackage{tikz} 

\usepackage{microtype}
\usepackage{graphicx}
\usepackage{subfigure}
\usepackage{booktabs} 
\usepackage{tikz}
\usepackage{algorithm}
\usepackage{algpseudocode}

\usepackage{hyperref}
\usetikzlibrary{fit}

\usepackage{amsmath}
\usepackage{amssymb}
\usepackage{mathtools}
\usepackage{amsthm}

\usepackage{graphicx} 
\usepackage{xcolor}    
\usepackage{caption,subcaption}

\usepackage{bm}
\usepackage{glossaries}
\usepackage[capitalize,noabbrev]{cleveref}
\usepackage{authblk}

\theoremstyle{plain}

\theoremstyle{definition}

\theoremstyle{remark}

\usepackage[textsize=tiny]{todonotes}
\usepackage{lipsum}

\title{Graph Agnostic Causal Bayesian Optimisation}

\author[*,1]{Sumantrak Mukherjee}
\author[*, 2]{Mengyan Zhang}
\author[2]{Seth Flaxman}
\author[1,3]{Sebastian J. Vollmer}
\affil[1]{%
    Department of Data Science and its Applications\\
    German Research Centre for Artificial Intelligence(DFKI)\\
}
\affil[2]{%
    Department of Computer Science\\
    University of Oxford\\
}
\affil[3]{%
    Department of Computer Science\\
    University of Kaiserslautern-Landau\\
}

\begin{document}
\maketitle

\begin{abstract}
We study the problem of globally optimising a target variable of an \textit{unknown causal graph} on which a sequence of soft or hard interventions can be performed. The problem of optimising the target variable associated with a causal graph is formalised as Causal Bayesian Optimisation (\textsc{cbo}), which has various applications in biology, manufacturing, and healthcare. However, in many real-world applications, the true data-generating process is often unknown or partially known.
We study the \textsc{cbo} problem with unknown causal graphs for two settings, namely structural causal models with hard interventions and function networks with soft interventions. We propose \textit{Graph Agnostic Causal Bayesian Optimisation} (\textsc{gacbo}), an algorithm that actively discovers the causal structure that contributes to achieving optimal rewards. \textsc{gacbo} seeks to balance exploiting the actions that give the best rewards against exploring the causal structures and functions.  
We show our proposed algorithm outperforms baselines in simulated experiments and real-world applications. 

\end{abstract}

\section{Introduction}

\textit{Bayesian Optimisation} (\textsc{bo}) is a powerful framework for sequentially optimising black-box functions, with various applications in fields such as drug and material discovery, robotics, agriculture, and automated machine learning \citep{movckus1975bayesian, garnett_bayesoptbook_2023}.
Although most conventional Bayesian optimisation methods \citep{srinivas2009gaussian,garnett_bayesoptbook_2023} treat functions as black boxes, the actual data-generating process usually exhibits some structural patterns, such as a network structure. 
Causal Bayesian Optimisation methods have been proposed to leverage underlying structures  
\citep{aglietti2020causal, sussex2022model} and
enable us to improve sample efficiency.

It is unrealistic to assume that the structural patterns are known in practice. In many real-world applications causal graphs are often unknown or wrongly specified (refer to Figure \ref{fig:IllustrativeDAGs} (c-w1), (c-w2) as examples). 
To solve this problem, we consider the Causal Bayesian Optimisation problem with an \emph{unknown causal graph}. 
One solution is to perform causal discovery (structure learning) \citep{glymour2019review} as a prior step to employing such techniques. 
However, learning the \textit{complete} true graph and all associated mechanisms \textit{globally} is not essential and possibly wasteful when the goal is finding the global optima. 
Previous works \citep{branchini2023causal,alabed2022bograph, toth2022active} proposed methods that integrate causal discovery and causal reasoning by jointly modelling the causal graphs and associated mechanisms for active learning tasks or minimising simple regret, under hard intervention settings. 
Different from their objective, we consider the Bayesian optimisation task with both soft and hard interventions, where our objective is to maximise the cumulative reward with the help of learning causal structures at the same time.  

\begin{figure}[ht]
    \centering
    \includegraphics[scale=0.35]{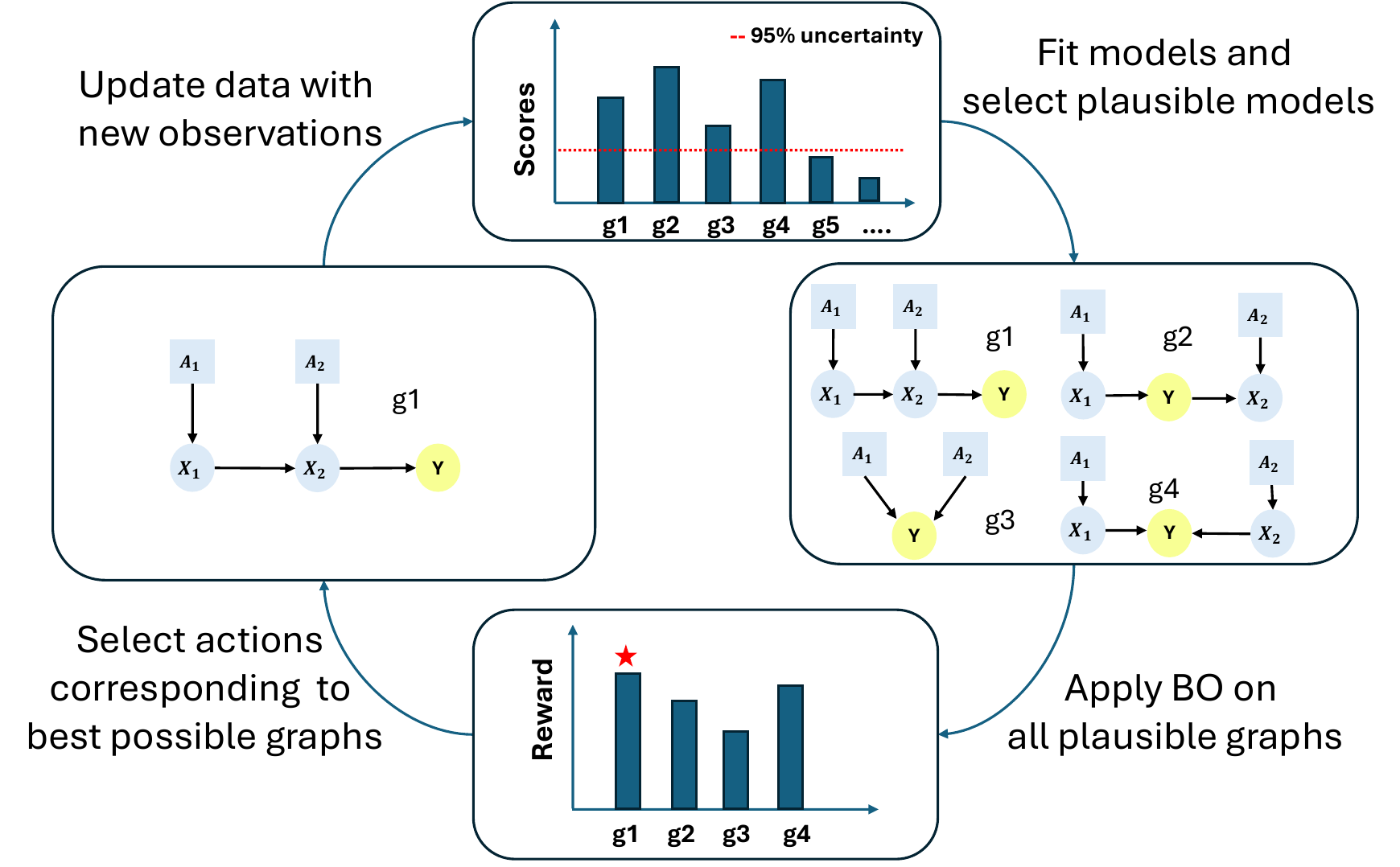}
    \caption{Graph Agnostic Causal Bayesian Optimisation (\textsc{gacbo}) workflow. \textit{Top}: Select plausible graphs based on data collected so far, \textit{Right}: Perform Causal Bayesian Optimisation on plausible graphs, \textit{Bottom}: Select the action based on the highest reward among all plausible graphs, \textit{Left}: Execute selected action, collect Data and repeat steps.}
\label{fig:workflow}
\end{figure}

    

To leverage graph structure for \textsc{bo} without wasting additional samples by learning inessential details of the graph structure,
we need to balance exploitation and exploration, where exploitation is in terms of selecting actions with the maximum possible outcome, and exploration is in terms of either selecting actions with high uncertainty in function space or causal structure learning.
We propose Graph Agnostic Causal Bayesian Optimisation (\textsc{gacbo}, Algorithm \ref{alg:gacbosoft} and \ref{alg:gacbohard}) with a causal subgraph discovery in Algorithm \ref{alg:Causal Subgraph Discovery}. \textsc{gacbo} optimises the target variable and learns the causal structure only when it serves the purpose of finding a possible optimal action. 
We show the workflow of our proposed method in Figure \ref{fig:workflow}. 
Starting from a uniform prior over all possible acyclic graph structures, we model surrogate functions for all ancestral nodes of the target for all possible graphs with Gaussian processes which take the parents and actions affecting the node as inputs. The probability of possible graphs is modelled using the Bayesian Score \citep{friedman2013gaussian}. 
In each round, we maintain the set of functions and graphs that are within specified confidence intervals with high probability.
We select interventions using an Upper Confidence Bound (\textsc{ucb}) based acquisition function utilising the reparametrization trick \citep{sussex2022model}. 

Our \textbf{contributions} are: 
\textbf{1)} To the best of our knowledge, we are the first work studying causal Bayesian optimisation with cumulative objective when the graph is unknown or partially known. 
\textbf{2)} We consider both soft and hard interventions and propose a novel algorithm Graph Agnostic Causal Bayesian Optimisation (\textsc{gacbo}) to address the new setting. 
\textbf{3)} We introduce an \textit{Upper Confidence Bound} based acquisition function that makes causal discovery a subtask only relevant when distinguishing between graphs yields better results, balancing the exploitation and exploration for causal structure learning. 
\textbf{4)} We demonstrate on synthetic and real-world causal graphs that our algorithm demonstrates competitive performance compared with baselines.


\section{Background \& Problem Setting}


In this paper, we address the novel and challenging setting where an agent interacting with a Structural Causal Model (\textsc{scm}) or a Noisy Function Network (\textsc{nfn}), with unknown graph structure and unknown associated functions, to maximise the cumulative rewards over $T$ rounds. We provide the background of \textsc{scm} and \textsc{nfn}, and define our problem setting.
We use uppercase letters to denote random variables and lowercase letters to denote the \textit{realisation} of random variables. We overload the notation of random variables to denote nodes in the causal graph. Bold letters denote vectors or sets.
We summarise our notation in Appendix \ref{sec:nomenclature}.

\begin{figure}
            \centering
        \includegraphics[scale=0.2]{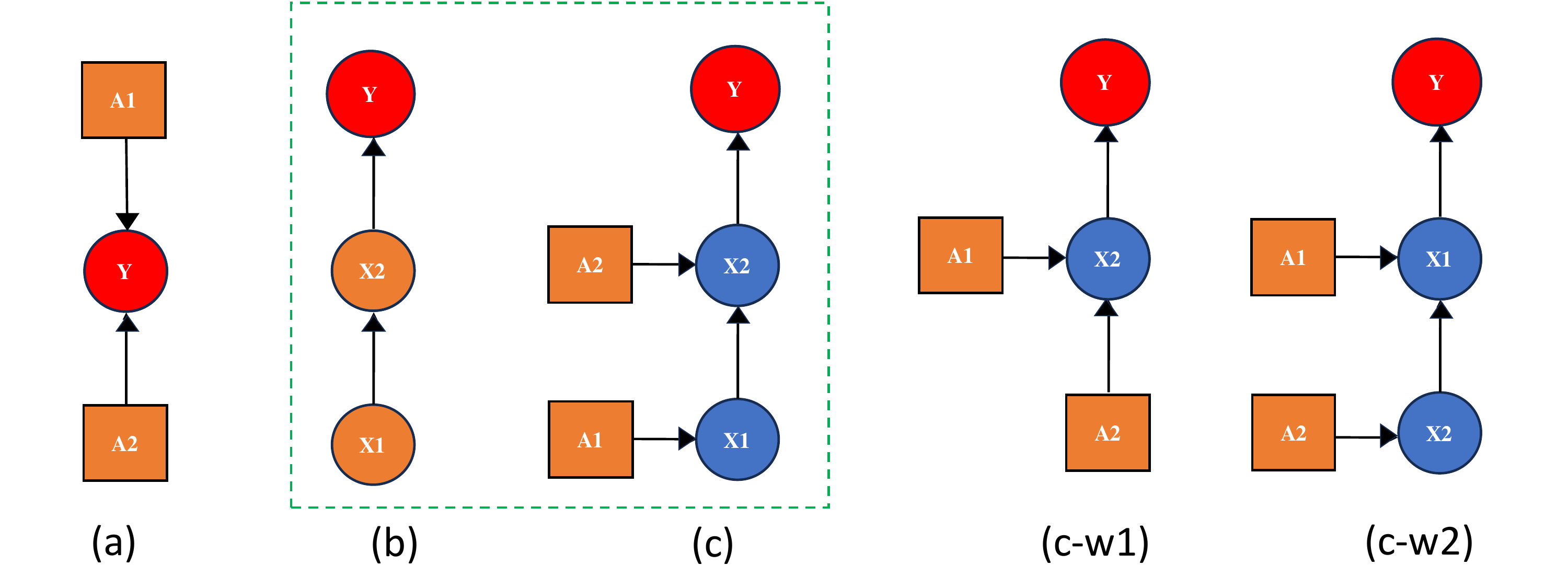}
        \caption{Problem Settings and Causal Structures: (a) Bayesian optimisation; (b) Structural causal models and hard interventions;
    (c) Function networks and soft interventions; 
    (c-w1) Incomplete graph for (c), missing X1; (c-w2) Incorrect graph for (c), reversing order of X1 and X2. The blue circles $X1$ and $X2$ represent non-manipulative variables, the orange squares $A1$ and $A2$ represent actions that can be taken, and $Y$ is the outcome of interest.
    }
    \label{fig:IllustrativeDAGs}
\end{figure}

\subsection{Structural Causal Models and Hard Interventions}
An \textsc{scm} \citep{pearl_2009} is expressed as a tuple 
$\langle{g^*},Y,\bm{V},\bm{f}_{g^*},\bm{\Omega}\rangle$ with the following elements:
${g^*}$ is the \textbf{true} but \textbf{unknown}. Denote $[m]$ as the set of integers $\{0\dots m \}$. Directed Acyclic Graph (\textsc{dag}), $Y$ denotes the reward variable,  $\bm{V}=\{V_i\}_{i=0}^{m-1}$ denotes a set of observed random variables (nodes in ${g^*}$), where all nodes $i\in[m]$ belong to a compact $\mathcal{V}_i\in\mathbb{R}$
and $\bm{f}_{g^*} =\{f_i^{g^*}\}_{i=0}^m$ is the set of respective \textbf{unknown} functions associated with $g^*$ with a set of independent noises $\boldsymbol{\Omega}=\{\Omega_i\}_{i=0}^m$ with zero mean and \textit{known} distribution.

Denote the indices of parent nodes of any node as $pa_{g^*}(i) \subset [m]$ is defined for the \textsc{dag} ${g^*}$. $\bm{Z}_{i}^{g^*} = \{V_j\}_{j \in pa_{g^*}(i)}$ are the parents of the $i^{th}$ node in the true graph ${g^*}$.
Node $V_i \in \bm{V}$ are generated according to the function 
$f_i^{g^*} : \mathcal{Z}_{i,{g^*}}\xrightarrow{}\mathcal{V}_i$.
We observe the noisy output of each function $v_i = f_i^{g^*}(\bm{z}_{i}^{g^*}) + \omega_i$.
The functions are evaluated topologically starting from root to leaf nodes according to ${g^*}$.

We consider a realistic setting where not all observable variables are intervenable \citep{lee2019structural}. 
Denote $\mathcal{I} \subset \{0, \dots, m-1\}$ as the indices of variables that allow to be intervened.
The set of observable variables can be then decomposed into two mutually exclusive sets $\bm{V} = \{\bm{C},\bm{X}\}$, where $\bm{X}=\{V_j\}_{j\in \mathcal{I}}$ are intervenable variables and $\bm{C}=\{V_j\}_{j \notin \mathcal{I}}$ are non-intervenable variables.  We assume the target variable $Y = V_m$ is non-intervenable.

For \textsc{scm}, we consider \textbf{hard interventions} also referred to as $do-$interventions. For hard interventions we are required to pick the indices we intervene on $\bm{I} \in \mathcal{P}(\mathcal{I})$ from the powerset of possible indices. We are also required to choose the corresponding values $\bm{a}_{\bm{I}}=\{a_i\}_{i\in\bm{I}}$ we intervene with. In our setting the space of action values possible for a particular node $i \in \bm{I}_t$ is given by the compact space $\mathcal{A}_i \in \mathbb{R} $. For a particular set of indices $\bm{I}$ the intervention space is denoted using $\mathcal{A}_{\bm{I}} = \{\mathcal{A}_i\}_{i \in \bm{I}}$ and therefore the total space of interventions is given by $\mathcal{A}_\mathcal{I} = \{\mathcal{A}_{\bm{I}}\}_{\bm{I} \in \mathcal{P}(\mathcal{I})}$, 
At round $t$, we select indices $\bm{I}_t \in \mathcal{P}(\mathcal{I})$ and set the corresponding nodes to specific values 
$\{do(x_{i,t} = a_{i,t})\ \forall {i\in \bm{I}_t}, {a_{i,t} \in \mathcal{A}_i} \}$
We then observe the values of all nodes, for $i \in [m]$,
\begin{equation}
v_{i,t} = 
\begin{cases}
  f_i^{g^*}(\bm{z}_{i,t}^{{g^*}}) + \omega_{i,t} & \text{if } i \notin \bm{I}_t\\
  a_{i,t} & \text{if } i \in \bm{I}_t\\
\end{cases}
\end{equation}

Note $f_i$ is dependent only on its ancestors in the mutated graph ${g^*_I}$ where the intervened nodes have no dependence on their parents.
We define the optimal intervention as the indices and values that maximise the target reward variable, i.e.
\begin{equation}
    \bm{I}^*, \bm{a}^*_{\bm{I}^*}= 
    {argmax}_{\bm{I} \in \mathcal{P}(\mathcal{I}), \bm{a} \in \mathcal{A}_{\bm{I}}}\mathbb{E}[Y|do(\bm{x}_{\bm{I}} = \bm{a})]
\end{equation}
For brevity we denote $\bm{a}^* = (\bm{I}^*, \bm{a}^*_{\bm{I}^*})\in \mathcal{A}_\mathcal{I}$.


\subsection{Function Networks and Soft Interventions}
The key difference between an \textsc{nfn} and \textsc{scm} is that, for \textsc{nfn} \citep{astudillo2021bayesian} one cannot directly set the value of a particular node in the function network. 
We consider a \textbf{soft intervention} model \citep{eberhardt2007interventions}, within this framework, interventions are described by controllable action variables.
These actions  $a_i$ belonging to a compact space $\mathcal{A}_i \in \mathbb{R}$ appear as additional nodes in the \textsc{nfn} with edges directed to the variable they affect (see Figure \ref{fig:IllustrativeDAGs}).

Since these actions affect nodes in conjugations with their existing inputs, we treat them as additional inputs, each node can be affected by multiple action variables, and the action affecting node $V_i$ is defined as
$\bm{a}_i^{g^*}=\{a_j\}_{j \in \bm{I}_i^{g*}}$ where $\bm{I}_i^{g*}$ is the set of actions affecting node $i$ in $g^*$. $\mathcal{A}_i^{g^*}\in \mathbb{R} ^q$ denotes the compact space of an input $\bm{a}_i^{g^*}$
where
($q=\arg\max_i|\bm{I}_i^{g*}|$ denotes the maximum number of actions affecting any node in the \textsc{nfn}). We define the total space of actions as $\bm{a} \in \mathcal{A}$ where $\bm{a}=\{\bm{a}_i^{g^*}\}_{i=0}^{m}$. Considering any node $V_i$ is dependent on the value of its parents $\bm{Z}^{g*}_i$ as well as the actions $\bm{a}_i^{g^*}$ affecting that node we define the set of functions as $\bm{f}_{g^*} = \{f_i^{g^*}\}_{i=0}^{m-1}$, with $f_i^{g^*}: \mathcal{Z}_i^{g^*} \times \mathcal{A}_i^{g^*} \xrightarrow{} \mathcal{V}_i$, with
$v_i^{g^*} = f_i^{g^*}(\bm{z}_i^{g^*}, \bm{a}_i^{g^*}) + \omega_i$.However, the functional relationships relating the actions to the nodes are unknown to us apriori.
We overload notations for \textsc{scm} and \textsc{nfn} cases for simplified notations, we will refer to different settings whenever needed. 

At round $t$, we select actions $\boldsymbol{a}_{:,t} = \{\bm{a}_{i,t}^{g^*}\}_{i=0}^{m-1}$ and observe nodes $\boldsymbol{v}_{:,t} = \{v_{i,t}\}_{i=0}^m$,
where $\boldsymbol{v}_{:,t}$ are generated by the oracle topologically according to the graph ${g^*}$, 
\begin{equation}
    v_{i,t} = f_i^{g^*}(\bm{z}_{i,t}^{{g^*}},\bm{a}_{i,t}^{g^*}) + \omega_{i,t},  \forall i \in [m]
\end{equation}
The root nodes in the graph have no observable parents but may still be affected by actions.
The target node has $\bm{a}_m=\{0\}_{j\in\bm{I}_m^{g^*}}$ since interventions on the target node are not allowed.
The value of the target variable depends on the actions of all variables that are ancestors of the target variable in graph ${g^*}$.
The action that maximises the target variable is defined using 
\begin{equation}\label{softoptimisation}
    \bm{a}^* = \arg\max_{\bm{a}\in\mathcal{A}}\mathbb{E}[y|\bm{a}]
\end{equation}

\paragraph{Performance Metric} Considering that the goal of our agent is to design a sequence of actions $\{\bm{a}_{:,t}\}_{t=0}^{T}$ or $\{\bm{I}_t,\bm{a}_{\bm{I}_t,}\}_{t=0}^{T}$ that maximises the average expected reward for soft and hard interventions respectively, which is equivalent as minimising the expected \textit{cumulative regret} \citep{sussex2022model, lattimore2020bandit},
\begin{equation}
\begin{aligned}
    R_T = \sum_{t=1}^T\left[\mathbb{E}[y|\bm{a}^*]-\mathbb{E}[y|\bm{a}_{:,t}]\right]; \quad
    R_T = \sum_{t=1}^T\left[\mathbb{E}[y|\bm{a}^*]-\mathbb{E}[y|do(\bm{x}_{\bm{I}_t}=\bm{a}_{\bm{I}_t})]\right].
\end{aligned}
\end{equation}
Iteratively evaluating the costly function at multiple points enhances the accuracy of surrogate estimates. 
If we assume that the true data generating mechanism complies with our regularity assumptions, the true data generating mechanism lies within our prior. Over time as we observe more data we can refine our estimate of the surrogate model.
Refining global estimates of function values reduces the occurrence of suboptimal actions, resulting in a sublinear growth of regret, which implies vanishing average regret i.e $R_T/T \xrightarrow[]{} 0$ as $T \xrightarrow[]{} \infty$.

\paragraph{Regularity Assumption} We operate under standard smoothness assumptions for any function relating any node to its parents $f_i^{g^*} \xrightarrow{}\mathcal{S} \times \mathcal{V}_i$ is defined over a compact domain $\mathcal{S}$.
For all nodes $i \in [m]$, we assume $f_i^{g^*}(\cdot)$ belongs to a reproducible Kernel Hilbert Space (RKHS) $\mathcal{H}_{k_i^{g^*}}$, a space of smooth functions defined on the input space $\mathcal{S} = \mathcal{Z}_i^{g^*} \times \mathcal{A}_i^{g^*}$ for \textsc{fn}s and $\mathcal{S} = \mathcal{Z}_i^{g^*}$ for \textsc{scm}s. 
This means all functions $f_i^{g^*} \in \mathcal{H}_{k_i^{g^*}}$ are induced by $k_i^{g^*} : \mathcal{S} \times \mathcal{S} \xrightarrow{} \mathbb{R}$. We also assume that $k_i^{g^*}(s,s') \leq 1$ for every $s,s' \in \mathcal{S}$. 
We enforce our smoothness assumptions by placing a bound on the RKHS norm of $f_i^{g^*}(\cdot)$, $\|f_i^{g^*}\| \leq \mathcal{B}_i$ for some fixed constant $\mathcal{B}_i \geq 0$.
To ensure the compactness of the domain $\mathcal{Z}_i^{g^*}$ we assume that the noise $\omega_i$ is either subgaussian or bounded i.e $\omega_i \in [-1,1]$.


\section{Method}

In this section, we propose the \textit{Graph Agnostic Causal Bayesian Optimisation (\textsc{gacbo})} algorithm to address the setting where both the causal graph and underlying causal mechanisms are unknown. 
By assuming the smoothness of functions relating the nodes and the presence of a causal structure in our data-generating process,  we can explore the plausible space of models more efficiently. 
There are three sources of uncertainty that we consider:\\
\textbf{1)} \textbf{Causal graph structure uncertainty.}
     We assume no prior knowledge of the \textsc{dag} structure and consider all \textsc{dags} equally likely. 
    To model this uncertainty, we place a uniform prior over all possible \textsc{dag}s and then update the posteriors over \textsc{dag}s as will be shown in Section \ref{sec: Plausible Models}. The updated beliefs of \textsc{dag}s enable us to update plausible graph sets and design acquisition functions.\\ 
 \textbf{2)} \textbf{Function uncertainty}. Given our regularity assumptions consider a large space of functions in the absence of data points. We use Gaussian Processes to model functions among nodes. This is the main type of uncertainty studied in standard BO techniques \citet{srinivas2009gaussian}, whereas in our setting due to the unknown graph structure, there are exponentially increasing numbers of functions with respect to nodes. We address function uncertainty in Section \ref{sec: Surrogate models}.\\
 \textbf{3)} \textbf{Noisy observations}. We consider noisy observations, however, we assume bounded variance or Subgaussian noise so that the input space $\mathcal{Z}_{i}^g$ is compact. We model this uncertainty in the \textsc{gp}s.

Our method quantifies all the sources of uncertainty by following a model-based approach.
We construct a confidence interval and consider graphs and functions within the confidence interval. 
For all graphs within the confidence interval, the observational uncertainty based on the confidence intervals of individual \textsc{gp}s is propagated through the graph structures. 
As we encounter data, the uncertainty reduces and our posterior converges to the true graph functions. 


\subsection{Surrogate models}
\label{sec: Surrogate models}
Surrogate models help us incorporate our prior beliefs into the modelling process and allow us to enact interventions without performing them in the real environment and also quantify the total uncertainty related to certain outcomes.
Define surrogate model $m_t \sim \mathcal{M}_t$ at time step $t$ as a triple $m_t = (g_t, \Tilde{\bm{f}}_t^g, \boldsymbol{\omega}_t^2)$.
$\mathcal{M}_t$ denotes the posterior of plausible models,
$g_t \sim G_t$ is one possible realisation of posterior $G_t$ at time $t$, 
$\Tilde{\bm{f}}_t^g = \{\Tilde{f}_{t,i}^g\}_{i=0}^m$ where the surrogate function $\Tilde{f}_{t,i}^g \in \mathcal{H}_{k_i^g}$ belongs to the RKHS $\mathcal{H}_{k_i^g}$ which is defined on the input space $\mathcal{S}_i^g = \mathcal{Z}_i^g \times \mathcal{A}_i^g$ (and $\mathcal{S}_i^g = \mathcal{Z}_i^g $ for hard interventions) for all nodes $i$ as implied by kernel $k_i^g : \mathcal{S}_i^g \times \mathcal{S}_i^g \xrightarrow[]{} \mathbb{R}$. 
And we assume subgaussian observational noise of each node $\boldsymbol{\omega}_t^2 = \{\omega_{t,i}^2\}_{i=0}^m$.

\paragraph{Surrogate Functions}We model surrogate functions using Gaussian processes (\textsc{gp}s). Posterior means $\mu_{i,t}^g$ and variances $\sigma_{i,t}^g$ for any function parameterising any possible graph $g$ at a given point is calculated according to the \textsc{gp} posterior at time step $t$. The posterior is calculated using \textsc{gp} update equations \citep{williams1995gaussian}. 
\begin{equation}\label{eq:gaussianprocess}
\Tilde{f}_{i,t}^g(\bm{z}_i^g,\bm{a}_i^g) \sim  \mathcal{N}(\mu_{i,t}^g(\bm{z}_i^g,\bm{a}_i^g),\sigma_{i,t}^g(\bm{z}_i^g,\bm{a}_i^g)),
\end{equation}
where
\begin{equation}\label{eq:gaussianprocessupdate}
\begin{aligned}
    \mu_{i,t}^g(\bm{z}_i^g,\bm{a}_i^g) &=\bm{k}_{i,t}^g(\bm{z}_i^g,\bm{a}_i^g)^{\top}(\bm{K}_{i}^g + \rho_i^2\bm{I})^{-1}\textrm{vec}(v_{i,1:t});\\
    \sigma_{i,t}^g(\bm{z}_i^g,\bm{a}_i^g) &= k_{i}^g((\bm{z}_i^g,\bm{a}_i^g),(\bm{z}_i^g,\bm{a}_i^g)) -
    \bm{k}_{i,t}^g(\bm{z}_i^g,\bm{a}_i^g)^{\top}(\bm{K}_{i}^g + \rho_i^2\bm{I})^{-1}\bm{k}_{i,t}^g(\bm{z}_i^g,\bm{a}_i^g),
\end{aligned}
\end{equation}
where $\bm{I}$ is used to define the identity matrix, vec$(v_{i,1:t}) = [v_{i,1} \dots v_{i,t}]^\top$ , 
$\bm{k}_{i,t}^g(\bm{z}_{i}^g,\bm{a}_i^g)= [k_{i}^g((\bm{z}_{i,1}^g,\bm{a}_{i,1}^g),(\bm{z}_{i}^g,\bm{a}_i^g)),
 \dots, k_{i}^g((\bm{z}_{i,t}^g,\bm{a}_{i,t}^g),(\bm{z}_{i}^g,\bm{a}_i^g))]$, 
 $[\bm{K}_{i}^{g}]_{t_1,t_2} = k_{i}^g((\bm{z}_{i,t_1}^g,\bm{a}_{i,t_1}^g),(\bm{z}_{i,t_2}^g,\bm{a}_{i,t_2}^g))$. 

\textbf{Graph Likelihood} 
The \textit{Markov Property} of Bayesian networks allows for a compact factorisation of the joint distribution of all observed nodes $\bm{V}= \{V_1,\dots, V_m\}$ in the Bayesian Network,
\begin{equation}
    p(\bm{V}|g) = \prod_{i=0}^m p(V_i|\bm{Z}_i^g).
\end{equation}
The joint distribution factorises into conditional distributions given its parents in the graph $g$.
In the case of soft interventions any observed node $V_i$ is affected by its parents $\bm{Z}_i^g$ as well as the actions which appear as extra nodes in the \textsc{scm}, we use $\bm{A}_i^g$ to denote the set of action nodes affecting node $i$ therefore it is calculated as
\begin{equation}
    p(\bm{V}|g) = \prod_{i=0}^m p(V_i|\bm{Z}_i^g,\bm{A}_i^g)
\end{equation}

The distribution factorises into conditional distributions for each variable, given its parents in the \textsc{dag} and the associated actions for the node.
\textsc{gp}s admit a closed-form expression for the marginal likelihood of the $t$ observations $v_{i,1:t}$ of the node $V_i$. $p(v_{i,1:t}|g, \bm{\theta}_i)$ can be calculated as below
\begin{equation}
\begin{aligned}
    (2\pi)^{-\frac{t}{2}}|\Tilde{\bm{K}}_{i,\theta}^g|^{-\frac{1}{2}}\textrm{exp}\left(-\frac{1}{2}v_{i,1:t}^{\top}(\Tilde{\bm{K}}_{i,\theta}^g)^{-1}v_{i,1:t}\right)
\end{aligned}
\end{equation}
where $\Tilde{\bm{K}}_{i,\theta}^g = \bm{K}_{i,\theta}^g + \omega_i^2I$. The covariance matrix $\bm{K}_{i,\theta}^g$ is given by the kernel $k_{i,\theta}^g$ used and observations collected until time step $t$ $(\bm{z}_{i,1}^g,\bm{a}_{i,1}^g) \dots (\bm{z}_{i,t}^g,\bm{a}_{i,t}^g) $,  $(\bm{z}_{i,1}^g) \dots (\bm{z}_{i,t}^g) $  for soft and hard interventions respectively.
The input space of the functions and hence the kernel specified is dependent on the selected graph. 
The lengthscales $\bm{\theta}_i=\{\theta_{i,j}\}_{i \in pa_g(i)}$ chosen for different input nodes in the selected graph, determine the smoothness of the functions in the \textsc{rkhs} implied by the kernel. 
The lengthscales chosen for the kernel relate directly to the smoothness of the functions sampled from the \textsc{gp} \citep{berkenkamp2019no}. We define priors $\bm{\theta}_i \sim \pi(\bm{\theta}_i)$ over hyperparameters consistent with our smoothness assumptions.

The \textit{Score} is defined as \citet{friedman2013gaussian} as $S$ and is calculated as follows.
The score shows the probability of the observed values of node $V_i$ is $v_{i,1:t}$ given the graph $g$ and dataset $\mathcal{D}_{t-1}$, where graph $g$ indicates the parents of node $V_i$ is $\bm{Z}_i^g$ and actions $\bm{A}_i^g$. 
\begin{equation}
    S(V_i, \bm{Z}_i^g, \bm{A}_i^g|\mathcal{D}_t)
    = \int p(v_{i,1:t}|g, \bm{\theta}_i) \pi(\bm{\theta}_i|g) d \bm{\theta}_i,
\end{equation}

Therefore the probability of observing data $\mathcal{D}_t$ given $g$ is given as the product of observing the values of each node in $i \in [m]$ given the values of its parents according to graph $g$ 
\begin{equation}
    P(\mathcal{D}_t | g) = \prod_{i=0}^m S(V_i, \bm{Z}_i^g, \bm{A}_i^g|\mathcal{D}_t).
\end{equation}
The probability of the graph $g$ given $\mathcal{D}_t$ is directly proportional to the product of the probability of observing the data given graph $ P(\mathcal{D}_t | g)$ and prior probability of graph $g$ $p(g)$ using Bayes Rule,
\begin{equation}
    P(g|\mathcal{D}_t) \propto P(\mathcal{D}_t|g)p(g).
\end{equation}




\subsection{Plausible Models} 
\label{sec: Plausible Models}
Define plausible models at time step $t$  as the set of surrogate models which is likely to contain the true \textsc{scm} within the confidence intervals with probability at least $1-\delta$
\begin{equation}\label{eq:plausiblenodes}
    |\Tilde{f}_{i,t}^g(\bm{z}_i^g,\bm{a}_i^g) -\mathbb{E}_{i,t}[\bm{z}_i,\bm{a}_i]| \leq \beta_{i,t}\sqrt{\mathbb{V}_{i,t}[\bm{z}_i,\bm{a}_i]},
\end{equation}
\begin{equation}
\begin{aligned}
    \mathbb{E}_{i,t}[\bm{z}_i,\bm{a}_i] &= \mathbb{E}_{g \sim p(g|\mathcal{D}_t)}[\mu_{i,t-1}^g(\bm{z}_i^g,\bm{a}_i^g)];\\
    \mathbb{V}_{i,t}[\bm{z}_i,\bm{a}_i] &= \mathbb{V}_{g \sim p(g|\mathcal{D}_t)}[\mu_{i,t-1}^g(\bm{z}_i^g,\bm{a}_i^g)]
    + \mathbb{E}_{g \sim p(g|\mathcal{D}_t)}[(\sigma_{i,t-1}^g(\bm{z}_i^g,\bm{a}_i^g))^2].
\end{aligned}   
\end{equation}
by the law of total expectation and law of total variance \citep{weiss2012elementary}.
Note that we use $\bm{z}_i$ instead of $\bm{z}_i^g$ as inputs for $\mathbb{E}_{i,t}(\cdot)$ and $\mathbb{V}_{i,t}(\cdot)$, in this case $z_i = [m]\backslash i$, where all observable nodes other than $V_i$ are inputs to the function.
The term $\beta_{i,t}$ is present to ensure confidence bounds. We set $\beta_{i,t} = \beta_{T}$ for all $i$. As shown in \citet{sussex2022model} some kernels exhibit dependence of $\beta_T$ on T, for $\beta_T$ to hold applying a union bound over all time steps $t$ and for all nodes $V_i$ for $i \in [m]$ is sufficient resulting in $\beta_T=\mathcal{O}(\mathcal{B}+\frac{\rho}{d}\sqrt{\gamma_t})$ where $\mathcal{B}=\max_i\mathcal{B}_i$ and $\rho=\max_i\rho_i$. 


At the time $t$ all $\Tilde{f}_i^g$ within the confidence intervals given by the joint posterior over graphs $p(g|\mathcal{D}_t)$ and associated GP posteriors constitute the \textit{plausible functions} $\mathcal{M}_t$.
\begin{equation}\label{eq:plausiblemodel}
    \begin{aligned}
    \mathcal{M}_t^g &= \{ 
        \Tilde{\bm{f}}_g = \{\Tilde{f}_i^g\}_{i=0}^{|V|} \text{ s.t. } \forall_i : \Tilde{f}_i^g \in \mathcal{H}_{k_i}, \| \Tilde{f}_i^g\|_{k_i} \leq \mathcal{B}_i, 
        \text{and } \eqref{eq:plausiblenodes} \textrm{ holds true } \forall a \in \mathcal{A}  
        \}.
    \end{aligned}
\end{equation}
Recall $\mathcal{B}_i$ is the upper bound of the RKHS norm of $f_i^{g^*}(\cdot)$, $\|f_i^{g^*}\| \leq \mathcal{B}_i$ for some fixed constant $\mathcal{B}_i \geq 0$.
The \textit{plausible graphs} $G_t$ are described as 
\begin{equation}
\label{equ: plausible graph}
    G_t = \{g | \forall_i \exists \Tilde{f}_i^g  \in \mathcal{M}_t^g,  g \in G_{t-1}\},
\end{equation}

with $G_0 = {g \in \mathcal{G}}$ where $\mathcal{G}$ denotes the space of all possible \textsc{dags}.
The above equation states that for a graph to be one of the plausible graphs at time $t$ there has to be at least one associated function with graph $g$ within the confidence interval for all nodes $[m]$.


\subsection{Acquisition Function}
\label{sec: Acquisition Function}
In addition to \textsc{mcbo} \citet{sussex2022model}, we design our acquisition function to incorporate uncertainty stemming from the absence of a priori knowledge regarding the underlying graph structure, by selecting actions giving the highest possible values according to the plausible graphs and functions,
\begin{equation}\label{eq:acquistionfn1}
    \bm{a}_{:,t} = \arg \max_{\bm{a}\in \mathcal{A}} \max_{g \in G_t}\max_{\Tilde{\bm{f}}_g\in \mathcal{M}_t^g} \mathbb{E}[y|\Tilde{\bm{f}}_g, \bm{a}].
\end{equation}
For a sampled graph $g$ nodes $V_i$ are evaluated topologically starting from root nodes $\{i \mid \bm{Z}_i^g = \emptyset\}$ in the graph $g$ to the target node $m$, based on actions $\bm{a}_{:,t}$.
For a given input to node $V_i$ a function $\Tilde{f}_i^g$ within the confidence interval is chosen optimistically such that it results in a desired input $\Tilde{v}_{i,t}^g$ for children nodes $\{j\mid i \in Z_j^g\}$ of $V_i$. Resulting in $\Tilde{\bm{v}}_t^g=\{\Tilde{v}_{i,t}^g\}_{i\in g}$ a hallucinated set of values for observed nodes included in graph $g$.

Optimising Eq.\eqref{eq:acquistionfn1} using standard techniques is not effective because it involves maximizing a set of functions characterized by bounded (RKHS) norms. To mitigate this problem we use the reparametrization trick used in \citet{sussex2022model} to write any function $\Tilde{f}_i^g \in \Tilde{\bm{f}}_g$ using a function $\eta_{i,g} : \mathcal{Z}_i^g \times \mathcal{A}_i^g \xrightarrow{} [-1,1]$, that is,
\begin{equation}
\begin{aligned}
    \Tilde{f}_{i,t}^g(\Tilde{\bm{z}}_i^g,\Tilde{\bm{a}}_{i}^g) = \mu_{i,t-1}^g(\Tilde{\bm{z}}_i^g,\Tilde{\bm{a}}_{i}^g) + \beta_t\sigma_{i,t-1}^g(\Tilde{\bm{z}}_i^g,\Tilde{\bm{a}}_{i}^g)\eta_{i,g}(\Tilde{\bm{z}}_i^g,\Tilde{\bm{a}}_{i}^g),
\end{aligned}
\end{equation}
where $\Tilde{\bm{z}}_{i}^g$ denotes the hallucinated values of $\bm{Z}_i^g$ for simulated action $\Tilde{\bm{a}}_{:,t}$, 
The $\eta_{i,g}$ function chooses plausible yet optimistic models based on the confidence bounds of functions given a graph $g$. The acquisition function can therefore be expressed in terms of $\bm{\eta_g}:\mathcal{Z}^g\times\mathcal{A}^g \xrightarrow{} [-1,1]^{|V(g)|}$, where $|V(g)|$ is the number of nodes in the graph $g$,
\begin{equation}\label{eq:acquisitionwitheta}
    \arg \max_{\bm{a}\in \mathcal{A}}\max_{g \in G_t}\max_{\bm{\eta}_g(\cdot)} \mathbb{E}[y|\Tilde{\bm{f}}_g, \bm{a}].
\end{equation} 
The acquisition for hard intervention (Eq. \ref{eq:acquisitionfunctionhard} in Appendix) is based on 
the notion of Minimal Intervention Sets (\textsc{mis}) \citep{lee2018structural}. For each plausible graph, we only compare interventions within the MIS of the given graph to find the intervention which maximises the surrogate model associated with that particular graph and then compare across all possible graphs to find the best plausible intervention. 

\begin{algorithm}[t]
\caption{Graph Agnostic Causal Bayesian Optimisation(Soft Interventions) (\textsc{gacbo-s})}
\label{alg:gacbosoft}
\begin{algorithmic}
\State {\bfseries Input:}  Parameters $\{\beta_t\}_{t\geq1}$, $\Omega$, generic kernel function $k_i$, prior over possible $\psi_{i,0}$ graph components, prior means $\mu_{i,0}=0 \forall i \in [m]$.
\For{$t=1 \dots T$}
    \State Construct confidence bounds for plausible functions $\mathcal{M}_t$ as in Eq. (\ref{eq:plausiblemodel}).
    \State Construct plausible graphs $G_t$ as in Eq. (\ref{equ: plausible graph}) using Algorithm \ref{alg:Causal Subgraph Discovery}.
    \State Select $\bm{a}_{:,t} \in \arg \max_{\bm{a}\in \mathcal{A}}\max_{g \in G_t}\max_{\bm{\eta}_g(\cdot)} \mathbb{E}[y|\Tilde{\bm{f}}_g, a]$ as in Eq. (\ref{eq:acquisitionwitheta}).
    \State Observe all nodes $\bm{v}_t$ and update $\mathcal{D}_t=\mathcal{D}_{t-1} \cup \{\bm{v}_{t},\bm{a}_{:,t}\}$.
    \State Update posterior $\{\{\mu_{i,t}^g(\cdot), \sigma_{i,t}^g(\cdot)\}_{i=0}^{m}\}_{g \in G}$.
\EndFor
\end{algorithmic}
\end{algorithm}

\subsection{Algorithm}
\label{sec: Algorithm}

With the surrogate models, plausible models, and acquisition function defined in the previous section, we now introduce our proposed algorithm to address the challenging setting where the causal graph and associated functions are unknown. 
The Graph Agnostic Causal BO approach for soft interventions is summarised in Algorithm \ref{alg:gacbosoft}, and the hard intervention version in Algorithm \ref{alg:gacbohard}. The main difference is we select nodes to intervene on via do-calculus and also choose the values to intervene with for hard intervention.
For graph discovery in both algorithms, we introduce a new Causal Subgraph Discovery method in Algorithm \ref{alg:Causal Subgraph Discovery}. We discuss the scalability of our approach in Appendix \ref{sec: Scalability}.

In Algorithm \ref{alg:gacbosoft}, we start by constructing confidence bounds for all plausible functions and update the plausible graph set as shown in Section \ref{sec: Plausible Models}. Before observing any data, all possible graphs are in the plausible graph set and have equal prior. 
Then for every feasible graph $g_t$ at time step $t$, we determine the action that maximizes the upper confidence bound of the variable of interest within that particular graph via the reparametrization trick as shown in Section \ref{sec: Acquisition Function}. We select the action corresponding to the plausible graph with the highest \textsc{ucb}.
After selecting the interventions, we can observe the corresponding values from all possible nodes and the dataset is updated which we use to update the GP posteriors for all plausible function. Based on the new posteriors we define new confidence bounds and repeat the steps for $T$ iterations.

For graph discovery in Algorithm \ref{alg:Causal Subgraph Discovery}, we recursively identify parent variables of observed nodes starting from the target node while ensuring acyclicity.
The algorithm initiates by computing the Gaussian Process Score $S(V_i, Z_i^g, A_i^g | \mathcal{D}_t)$ for all potential components, considering the input space composed of combinations of observed nodes and action variables $\mathcal{Z}_i \times \mathcal{A}_i$ across all nodes $i$.
 A list of descendants is maintained to guide the traversal, and while choosing the component for a node, components with inputs containing descendants of the node are eliminated. Normalization is then applied to the Scores of the remaining components, and sampling from a multinomial distribution based on normalized probabilities is performed. The discovery process introduces a bias based on the encountered nodes during traversal, as the descendants of a node are determined by the order of traversal. To avoid such a bias we randomly sample a permutation from a uniform distribution over all possible permutations, to choose the order of iteration. 
We repeat $n$ times to get $n$ graph samples.

Similar to the acquisition function defined in \ref{eq:acquistionfn1}, we define an acquisition function for hard interventions, with the only difference being that hard interventions are performed instead of soft interventions. The observational uncertainty is propagated through the resulting mutated graph, the reparameterisation trick is used to find optimistic upper confidence for all plausible graphs 
\begin{equation}\label{eq:acquisitionfunctionhard}
  \arg \max_{I,\bm{a}_I \in \mathcal{A}}\max_{g \in G_t}\max_{\bm{\eta}_g(\cdot)} \mathbb{E}[y|\Tilde{\bm{f}}_g, do(V_I = \bm{a}_I)].
\end{equation}

This is slightly different as compared to soft interventions, because a hard intervention mutates the graph, making the node independent of all ancestor nodes and interventions performed on them, thus simplifying the problem.
This induces the notion of Minimal Intervention Sets \textsc{mis} \citep{lee2018structural}.
A \textsc{mis} for an \textsc{scm} $\langle{g},Y,\bm{V},\bm{f}_{g},\bm{\Omega}\rangle$ is defined as the set of variables $\bf{X}_s \in \mathcal{P}(\bf{X})$ such that there exists no such $\bf{X}^{'}_s \subset \bf{X}_s$ for which $\mathbb{E}[Y\mid do(\bf{X}_s)] = \mathbb{E}[Y\mid do(\bf{X}'_s)]$. We denote the \textsc{mis} for graph $g$ with target node $y$ as $\mathbb{M}_{g,y}$
 however since the graph structure is not known to us a priori, we construct our \textit{Plausible MIS} $\mathbb{M}_{y,t}$, by taking the union over the \textsc{mis} of plausible graphs at time step $t$, i.e.
$\mathbb{M}_{y,t} = \bigcup_{g \in G_t} {\mathbb{M}_{g,y}}$.
For each plausible graph, we only compare interventions within the MIS of the given graph to find the intervention which maximises the surrogate model associated with that particular graph and then compare across all possible graphs to find the best plausible intervention. 

\begin{algorithm}[t]
\caption{Graph Agnostic Causal Bayesian Optimisation (Hard intervention) (\textsc{gacbo-h})  }\label{alg:gacbohard}
\begin{algorithmic}
\State {\bfseries Requires:}  Parameters $\{\beta_t\}_{t\geq1}$, $\Omega$, generic kernel function $k_i$, prior over possible graphs $G_0$, prior means $\mu_{i,0}^g=0 \forall i \in [m], g \in G_0$.
\For{$t=1 \dots T$}
    \State Construct confidence bounds for plausible functions $\mathcal{M}_t$ as in \ref{eq:plausiblemodel}
    \State Construct plausible graphs as in \ref{equ: plausible graph}.

    \State Select $I,\bm{a}_I \in \arg \max_{I,\bm{a}_I \in \mathcal{A}}\max_{g \in G_t}\max_{\bm{\eta}_g(\cdot)}$ $\mathbb{E}.[y|\Tilde{\bm{f}}_g, do(V_I = \bm{a}_I)]$ as in \ref{eq:acquisitionfunctionhard}.
    \State Observe all nodes $\bm{v}_t$ and update $\mathcal{D}_t=\mathcal{D}_{t-1} \cup \{\bm{v}_{t},\bm{a}_{t}\}$
    Update posterior $\{\{\mu_{i,t}^g(\cdot), \sigma_{i,t}^g(\cdot)\}_{i=0}^{m}\}_{g \in G}$.
\EndFor
\end{algorithmic}
\end{algorithm}

\begin{algorithm}[t]
\caption{Causal Subgraph Discovery}\label{alg:Causal Subgraph Discovery}
\begin{algorithmic}
\State {\bfseries Input:} 
$\bm{S}_i=\{S(V_i, Z_i, A_i|\mathcal{D}_t), \forall  (Z_i, A_i)\in \mathcal{Z}_i \times \mathcal{A}_i$\}, where $i \in [m]$, $g_t = \{\}$, $De(m) = \{\}$.

\Function{FindSubgraph}{$g, i,\bm{S}_i,De(i)$}
    \If{$i \in g$}
        \State\Return $g$
    \EndIf
    \State $\bm{S}_i^c = \{S \in \bm{S}_i \mid Z_i \cap De(i) = \emptyset\}$,$\Tilde{\bm{S}}_i^c = \{\frac{S}{\sum_{S\in \bm{S}_i^c}^{}S} \forall S \in \bm{S}_i^c\}$,
    $(Z_i^c,A_i^c) \sim \text{Multinomial}(\Tilde{\bm{S}}_i^c)$
    \If{$Z_i^c = \emptyset$}
    \State $g = g\cup\{i:(Z_i^c,A_i^c)\}$
    \Else
    \State $g = \{g\cup\{i:(Z_i^c,A_i^c)\}$,
    $Pa_g(i) \sim \text{Uniform(Permutations}({Z_i^c}))$
    \For{$j \in Pa_g(i)$}
        \State $De(j)=De(j)\cup \{i\}$,
        $g = \Call{FindSubgraph}{g,j,\bm{S}_j, De(j)}$
    \EndFor
    \EndIf
    \State{\Return g}
\EndFunction{}
\State \textbf{Output:} $g_t = \Call{FindSubgraph}{g_t, m, \bm{S}_m, De(m)}$

\end{algorithmic}
\end{algorithm}

\section{Experiments}
\label{sec: experiments}



We evaluate 
our proposed algorithm \textsc{gacbo}
on synthetic environments (Dropwave, Alpine3, Rosenbrock, ToyGraph) and a real-life (Epidemiology Graph) environment introduced in \citep{astudillo2021bayesian, branchini2023causal}.
Our reported metric is average reward which is directly inversely related to cumulative regret. A higher expected reward indicates lower cumulative regret, which suggests better performance.
We repeat each experiment 5 times with different seeds and report average rewards $\pm \sigma/ \sqrt{5}$, where $\sigma$ is the standard deviation across all repeats.

\begin{figure}[t!]
    \centering
    \includegraphics[scale=0.25]{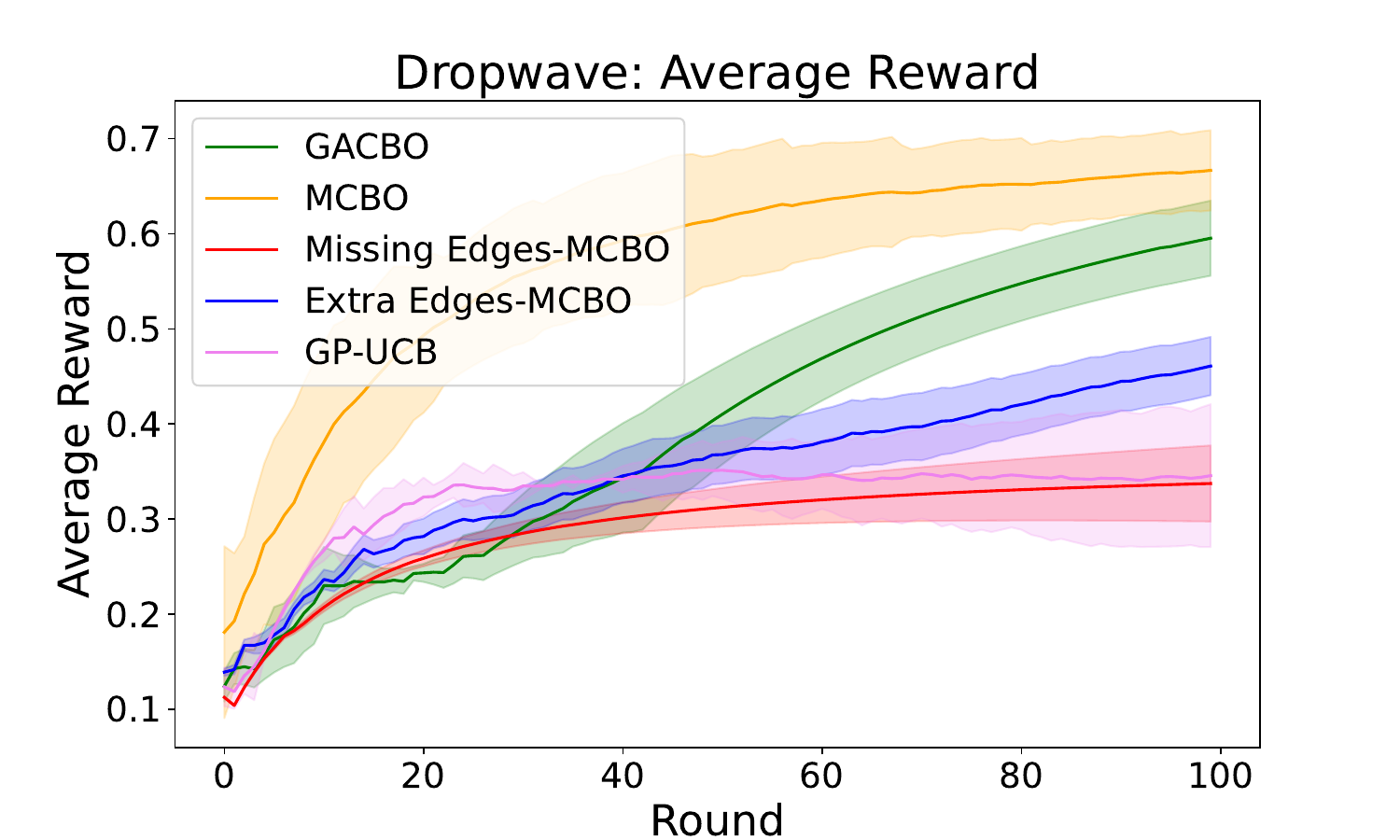}
    \includegraphics[scale=0.25]{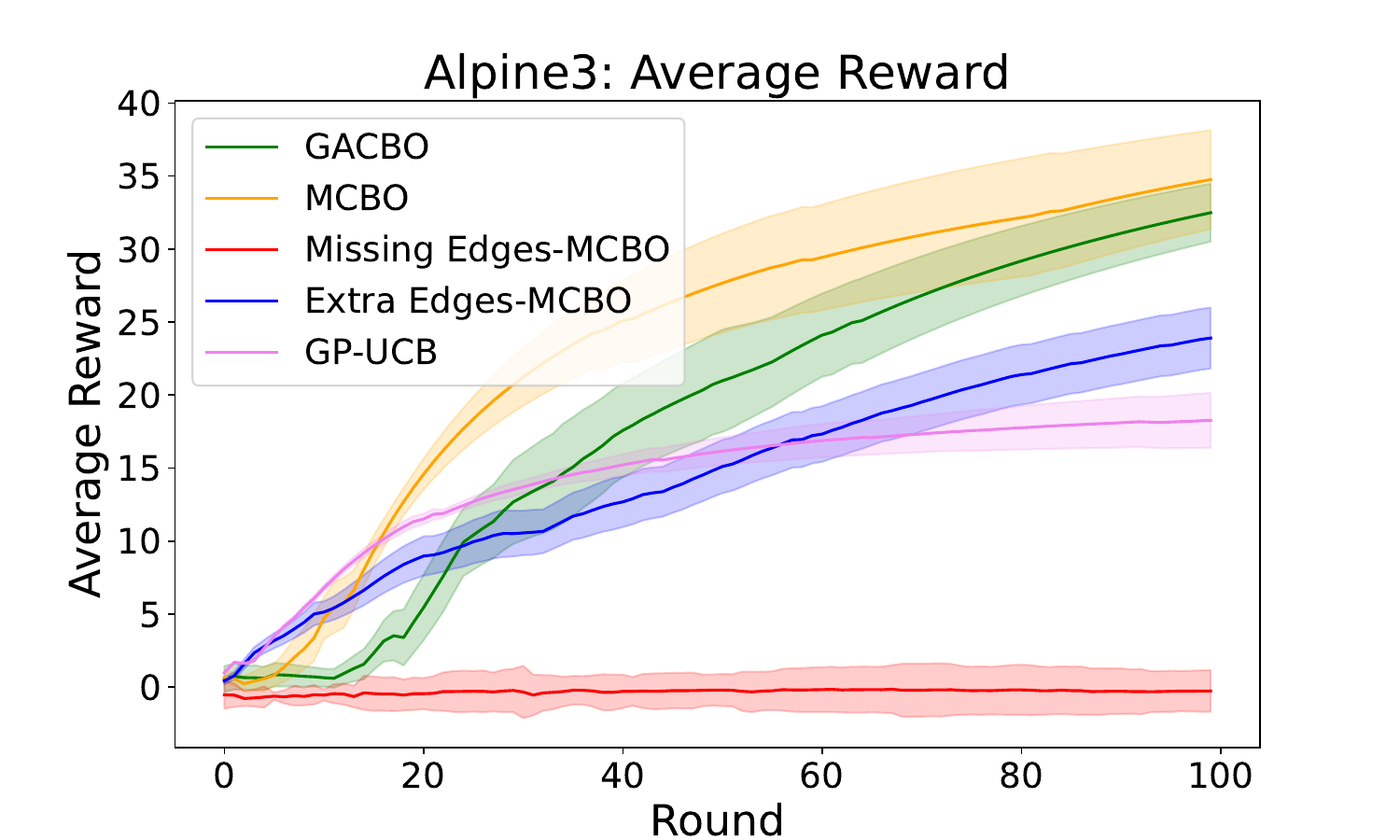}
    \includegraphics[scale=0.25]{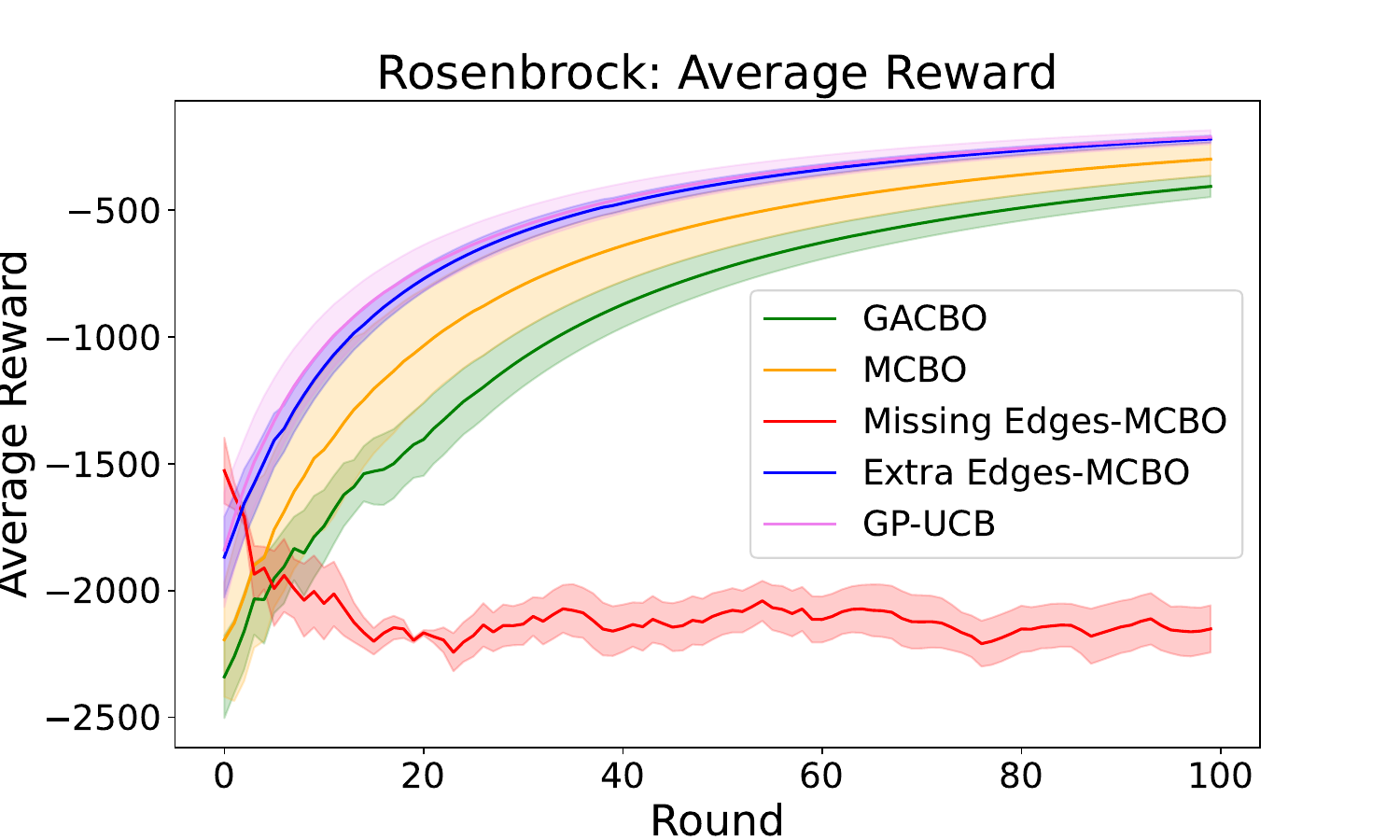}
    \includegraphics[scale=0.25]{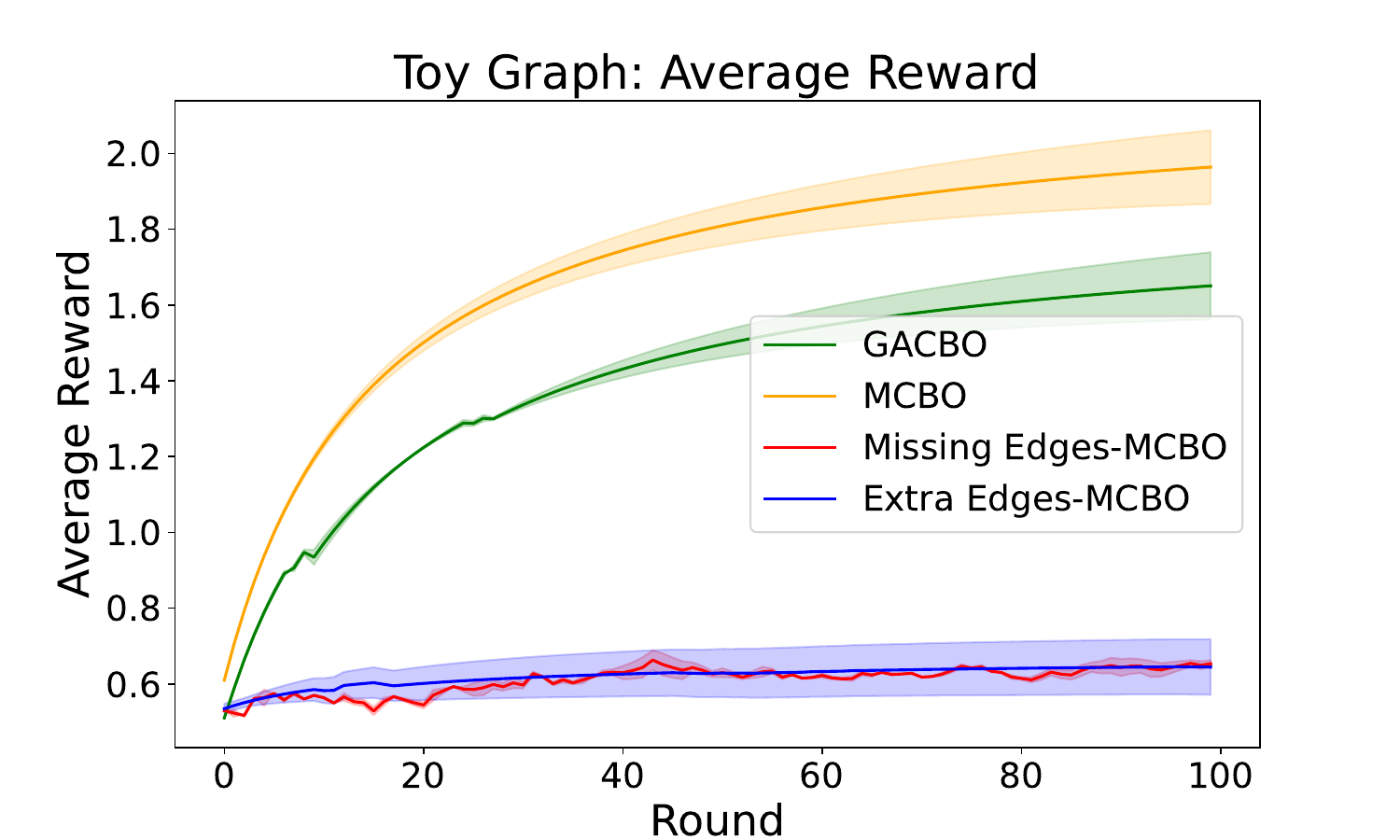}
    \caption{Simulation Results. We compare our method (\textsc{gacbo}) with \textsc{mcbo} with true graph and wrong graphs (missing edges and extra edges), and \textsc{gp-ucb}. As discussed in Section \ref{sec: experiments}, \textsc{gp-ucb} is inappropriate for ToyGraph. For soft intervention, we tested on Dropwave, Alpine3 and Rosenbrock. For hard intervention, we tested on ToyGraph.}
    \label{fig:SimulationResults} 
\end{figure}

\paragraph{Baselines}
We consider the following baselines to compare with our proposed algorithm
1) Model-based Causal Bayesian Optimisation \textsc{mcbo} \citep{sussex2022model} with the true causal graph, which is the state-of-the-art \textsc{cbo} method for cumulative regret objective.
2) We further test \textsc{mcbo} with incorrect graphs to simulate the situation where a causal graph might not be given or wrongly specified. We consider two kinds of incorrect graphs 1) Missing Edges: In such graphs at least one ancestral node of the target variable has a missing parent 2) Extra Edges: All true edges are present and additional edges are added. The correct graph and incorrect graphs used for experiments are provided in the appendix \ref{sec:SimulationDetails}
3) For soft intervention, we also compare with the BO algorithm Gaussian Process Upper Confidence Bound (\textsc{gp-ucb}) \citep{srinivas2009gaussian}, which indicates the performance without exploring and learning causal structure. 
\textsc{gp-ucb} cannot be compared with our algorithm in the hard intervention case, since hard interventions require choosing nodes to intervene on, and \textsc{gp-ucb} cannot handle that without modification. 

\paragraph{Simulations}
We show the performance of our simulated experiments in Figure \ref{fig:SimulationResults}.
For soft intervention, we consider three simulated graphs \citep{astudillo2021bayesian}, namely Dropware, Alpine3 and Rosebrock. 
For hard intervention, we consider the ToyGraph \citep{aglietti2020causal}, with the intervention sets $\mathcal{I}=\{\emptyset, \{0\},\{1\},\{0,1\}\}$.
The causal graphs we use can be found Figure \ref{fig:Toy Graph, correct and wrong} in the Appendix. 
As expected, with the true causal graph, \textsc{mcbo} shows the best performance for almost all experiments and indicates the average rewards we can get when we do not spend samples exploring causal structure. 
The only exception is Rosenbrock, \textsc{gp-ucb} outperforms \textsc{mcbo}, this is because decomposing the function based on the causal graph of Rosenbrock provides no significant advantage as the output of each function is simply added to the input of the following node.
We can observe that \textsc{mcbo} failed to learn good interventions and get good average rewards when there is no true causal graph revealing, even when the causal graph is slightly wrongly specified. When extra edges are specified the rate of learning a good intervention is significantly slower since the dimensionality increases. For the case of missing edges, the true function lies outside the function space and decisions are made on an incorrectly specified model which results in poor performance.
Our algorithm \textsc{gacbo} initially suffers a higher regret as it has no causal structure information and performs actions based on incorrect graph structures before it converges to the true graph. 
We observe it can learn the causal graph quickly and reach to similar performance as \textsc{mcbo} after around 100 rounds for all experiments. 
Our simulated experiments show that our proposed algorithm can efficiently learn causal structures which benefits the goal of maximising average (cumulative) rewards.

\begin{figure}
    \centering
    \includegraphics[scale=0.2]{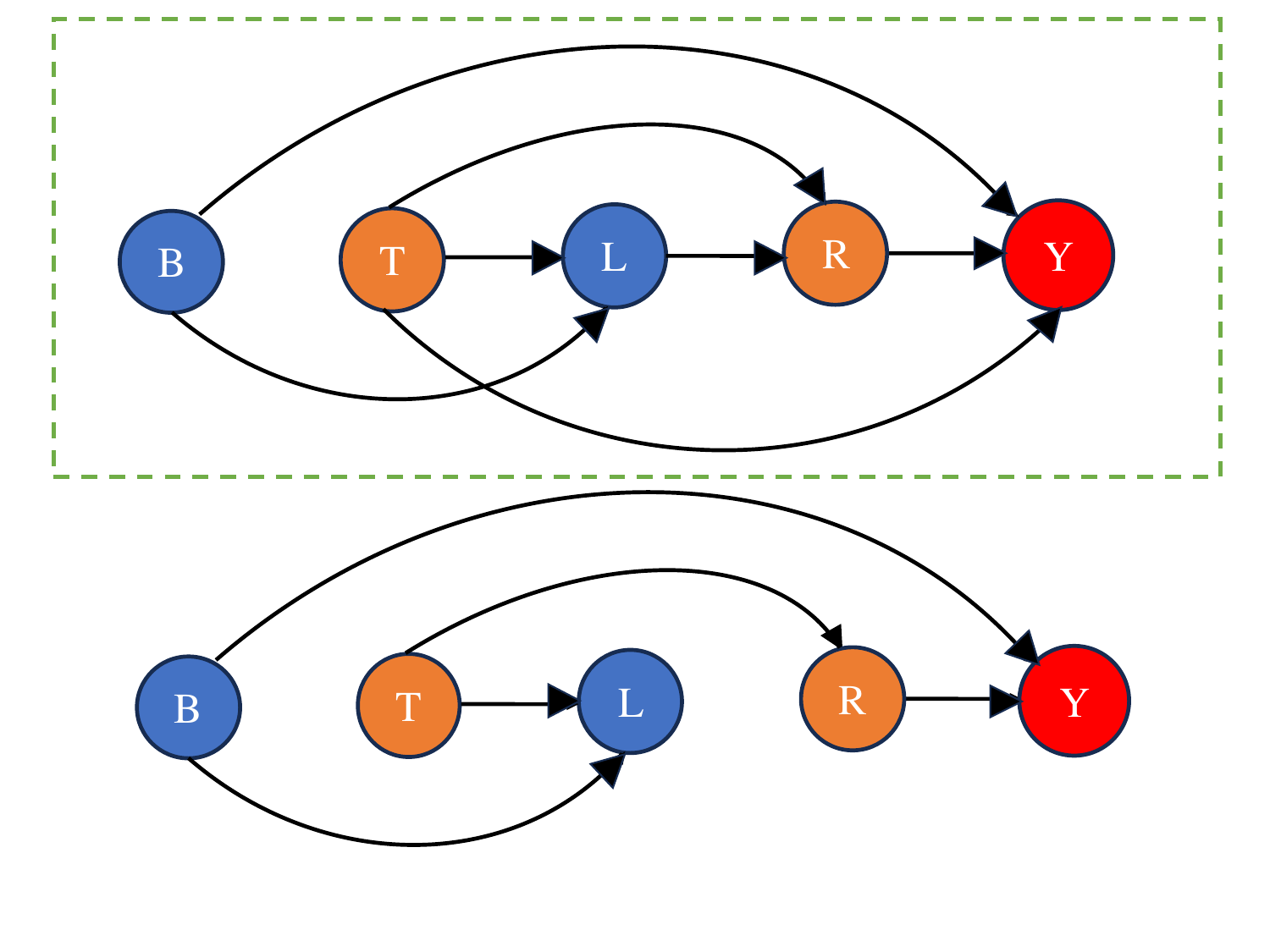}
    \includegraphics[scale=0.25]{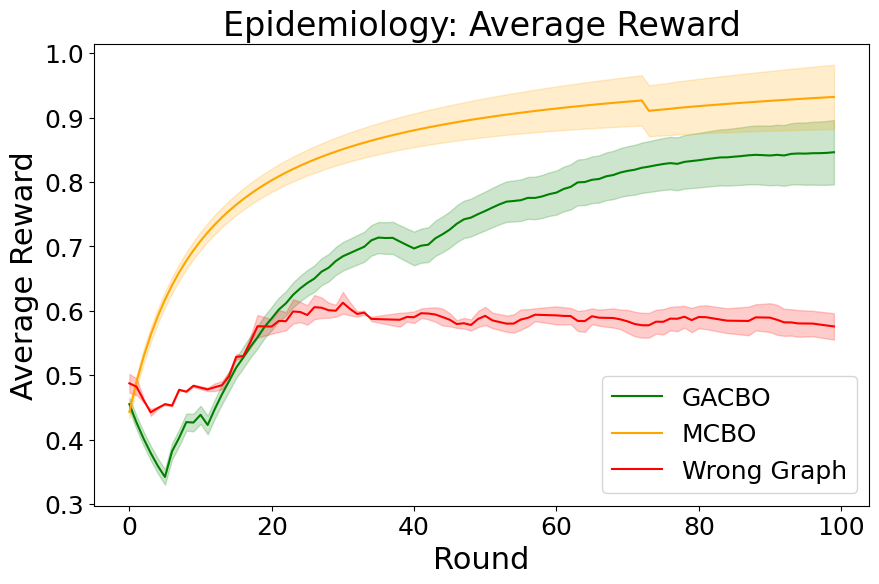}
    \caption{Real-world applications: Epidemiology. We show the true causal graph on the left top, where the T,R are potential treatments and Y is the target to be optimised. B,L are non-manipulative. The left bottom shows the designed wrong causal structure for \textsc{mcbo}. The right-hand side shows the performance of our experiments.}
    \label{fig:RealWorldSimulation}
\end{figure}

\paragraph{Real-World Application: Epidemiology}

We test our proposed algorithm in a real-world application in Epidemiology \citep{havercroft2012simulating, branchini2023causal}.
We show the associated causal graph and performance in Figure \ref{fig:RealWorldSimulation}.
The experiment aims to minimise HIV viral load by selecting two possible treatments (T, R, refer to \citet{havercroft2012simulating} for details).
The agent is allowed to perform two treatments referenced as T, R, and a combination of both treatments simultaneously, i.e., the possible intervention set is $\mathcal{I}=\{\emptyset,\{T\},\{R\},\{T,R\}\}$. 


This environment is significantly more complex than the ToyGraph Environment due to a multimodal objective function with high observational noise. 
We observe in this real-world example, that our algorithm significantly outperforms \textsc{mcbo} with a wrongly specified causal graph, even if the causal graph only has two missing edges. 
\textsc{gacbo} can learn the causal structure efficiently and achieve a similar level of performance compared with \textsc{mcbo} with a known true graph within 100 rounds. 



\section{Related Work}



\textbf{Causal Decision Making}
The first causal Bayesian optimisation setting was proposed in \citet{aglietti2020causal}, which focused on hard interventions and the best intervention identification setting. 
\citet{sussex2022model} expanded their setting to include soft interventions and noisy environments. They proposed the Model-based Causal Bayesian Optimisation (\textsc{mcbo}) algorithm, which is the state-of-the-art method with a known graph. 
With unknown graphs for cumulative regret objective, \citet{lu2021causal, de2022causal, konobeev2023causal} considered causal multi-armed bandits. \citet{lu2021causal} studied causal trees, causal forests and proper interval graphs, with regret analysis under a few causal assumptions. \citet{de2022causal} utilised an estimator based on separating sets, with no theoretical analysis on regret shown. 
\citet{konobeev2023causal} proposed a RAndomized Parent Search algorithm (RAPS) and showed conditional regret upper bounds.
\citet{malek2023additive} show that the unknown causal graph be exponentially hard in parents of the outcome and studies the problem under the additive assumption on the outcome.
All the above work considered discrete arms (intervention values) and linear bandits, while our work addresses continuous intervention values, non-linear relations between nodes and a more general class of graphs.

\citet{branchini2023causal} studied the \textsc{cbo} setting with an unknown graph for the best intervention identification setting.  Their approaches are based on the entropy search criterion. 
However, directly applying their method to cumulative regret objective would lead to suboptimal performance since one needs to further balance the exploitation-exploration balance between picking actions that lead to the best rewards and learning causal structures. 
\citet{Alabed2022} studied the \textsc{cbo} problem for unknown causal graph scenario with a specific application to autotuners. 
To the best of our knowledge, we are the first to study the \textsc{cbo} with unknown graph and cumulative regret objectives.

\paragraph{Active Causal Discovery} 
\citet{von2019optimal} developed a Bayesian optimal experimental design framework to perform active causal discovery for Gaussian Process networks. \citet{lorch2021dibs,giudice2023bayesian} addressed the problem of causal discovery for graphs with a larger number of nodes. Based on which, \citet{tigas2022interventions,tigas2023differentiable} performed active causal discovery for larger graphs. 
\citet{toth2022active} considered the active learning methods for unifying sequential causal discovery and causal reasoning. 

The goals of active causal discovery and Bayesian optimisation are misaligned. While Bayesian optimisation tries to balance exploration and exploitation to minimise cumulative regret, the active causal discovery acquisition function might choose an intervention that has a low reward and does not help future steps of CBO but helps discover the true underlying Causal Graph.  
Therefore it is sub-optimal to first perform active causal discovery and then followed by causal Bayesian optimisation as separate steps. 
Our algorithm naturally unifies these two steps by making causal discovery a sub-task of causal Bayesian optimisation. See Appendix \ref{sec: ACD - CBO} for a detailed discussion.

\section{Conclusion}

In this paper, we study a novel setting of causal Bayesian optimisation with the cumulative regret objective where the causal graph is unknown. To the best of our knowledge, our work is the first work addressing this setting.
This is motivated by real-world applications such as Epidemiology where we do not have prior knowledge of causal structure while we deal with optimisation problems.
For two settings, namely structural causal models with hard interventions and function networks with soft interventions, we propose a new algorithm Graph Agnostic Causal Bayesian Optimisation (\textsc{gacbo}) to balance picking actions with known high rewards and learning the causal structure and reducing function uncertainty by sampling unknown actions only when it potentially results in better rewards, based on the Upper Confidence Bound (\textsc{ucb}) type of acquisition function. Our algorithm deals with three types of uncertainties causal structure, function space, and noisy observations.
We show in simulations and real-world applications that our algorithm outperforms the state-of-art \textsc{mcbo} algorithm when there is no underlying true graph revealed to it. We discuss the theoretical analysis of our algorithm in Appendix \ref{sec: Discussion on Theoretical Analysis} and leave it as future work.

\newpage



\bibliographystyle{plainnat}
\bibliography{gacbo.bib}

\begin{thebibliography}{52}
\providecommand{\natexlab}[1]{#1}
\providecommand{\url}[1]{\texttt{#1}}
\expandafter\ifx\csname urlstyle\endcsname\relax
  \providecommand{\doi}[1]{doi: #1}\else
  \providecommand{\doi}{doi: \begingroup \urlstyle{rm}\Url}\fi

\bibitem[Aglietti et~al.(2020)Aglietti, Lu, Paleyes, and Gonz{\'a}lez]{aglietti2020causal}
Virginia Aglietti, Xiaoyu Lu, Andrei Paleyes, and Javier Gonz{\'a}lez.
\newblock Causal bayesian optimization.
\newblock In \emph{International Conference on Artificial Intelligence and Statistics}, pages 3155--3164. PMLR, 2020.

\bibitem[Agrawal et~al.(2019)Agrawal, Squires, Yang, Shanmugam, and Uhler]{agrawal2019abcd}
Raj Agrawal, Chandler Squires, Karren Yang, Karthikeyan Shanmugam, and Caroline Uhler.
\newblock Abcd-strategy: Budgeted experimental design for targeted causal structure discovery.
\newblock In \emph{The 22nd International Conference on Artificial Intelligence and Statistics}, pages 3400--3409. PMLR, 2019.

\bibitem[Alabed and Yoneki(2022{\natexlab{a}})]{Alabed2022}
Sami Alabed and Eiko Yoneki.
\newblock {BoGraph}.
\newblock In \emph{Proceedings of the 2nd European Workshop on Machine Learning and Systems}. {ACM}, April 2022{\natexlab{a}}.
\newblock \doi{10.1145/3517207.3526977}.
\newblock URL \url{https://doi.org/10.1145/3517207.3526977}.

\bibitem[Alabed and Yoneki(2022{\natexlab{b}})]{alabed2022bograph}
Sami Alabed and Eiko Yoneki.
\newblock Bograph: structured bayesian optimization from logs for expensive systems with many parameters.
\newblock In \emph{Proceedings of the 2nd European Workshop on Machine Learning and Systems}, pages 45--53, 2022{\natexlab{b}}.

\bibitem[Andersson et~al.(1997)Andersson, Madigan, and Perlman]{andersson1997characterization}
Steen~A Andersson, David Madigan, and Michael~D Perlman.
\newblock A characterization of markov equivalence classes for acyclic digraphs.
\newblock \emph{The Annals of Statistics}, 25\penalty0 (2):\penalty0 505--541, 1997.

\bibitem[Astudillo and Frazier(2021)]{astudillo2021bayesian}
Raul Astudillo and Peter Frazier.
\newblock Bayesian optimization of function networks.
\newblock \emph{Advances in neural information processing systems}, 34:\penalty0 14463--14475, 2021.

\bibitem[Berkenkamp et~al.(2019)Berkenkamp, Schoellig, and Krause]{berkenkamp2019no}
Felix Berkenkamp, Angela~P Schoellig, and Andreas Krause.
\newblock No-regret bayesian optimization with unknown hyperparameters.
\newblock \emph{arXiv preprint arXiv:1901.03357}, 2019.

\bibitem[Branchini et~al.(2023)Branchini, Aglietti, Dhir, and Damoulas]{branchini2023causal}
Nicola Branchini, Virginia Aglietti, Neil Dhir, and Theodoros Damoulas.
\newblock Causal entropy optimization.
\newblock In \emph{International Conference on Artificial Intelligence and Statistics}, pages 8586--8605. PMLR, 2023.

\bibitem[Chickering(2002)]{chickering2002optimal}
David~Maxwell Chickering.
\newblock Optimal structure identification with greedy search.
\newblock \emph{Journal of machine learning research}, 3\penalty0 (Nov):\penalty0 507--554, 2002.

\bibitem[De~Kroon et~al.(2022)De~Kroon, Mooij, and Belgrave]{de2022causal}
Arnoud De~Kroon, Joris Mooij, and Danielle Belgrave.
\newblock Causal bandits without prior knowledge using separating sets.
\newblock In \emph{Conference on Causal Learning and Reasoning}, pages 407--427. PMLR, 2022.

\bibitem[Eaton and Murphy(2007)]{eaton2007exact}
Daniel Eaton and Kevin Murphy.
\newblock Exact bayesian structure learning from uncertain interventions.
\newblock In \emph{Artificial intelligence and statistics}, pages 107--114. PMLR, 2007.

\bibitem[Eberhardt and Scheines(2007)]{eberhardt2007interventions}
Frederick Eberhardt and Richard Scheines.
\newblock Interventions and causal inference.
\newblock \emph{Philosophy of science}, 74\penalty0 (5):\penalty0 981--995, 2007.

\bibitem[Faria et~al.(2022)Faria, Martins, and Figueiredo]{faria2022differentiable}
Gon{\c{c}}alo Rui~Alves Faria, Andre Martins, and M{\'a}rio~AT Figueiredo.
\newblock Differentiable causal discovery under latent interventions.
\newblock In \emph{Conference on Causal Learning and Reasoning}, pages 253--274. PMLR, 2022.

\bibitem[Friedman and Koller(2003)]{friedman2003being}
Nir Friedman and Daphne Koller.
\newblock Being bayesian about network structure. a bayesian approach to structure discovery in bayesian networks.
\newblock \emph{Machine learning}, 50:\penalty0 95--125, 2003.

\bibitem[Friedman and Nachman(2013)]{friedman2013gaussian}
Nir Friedman and Iftach Nachman.
\newblock Gaussian process networks.
\newblock \emph{arXiv preprint arXiv:1301.3857}, 2013.

\bibitem[Garnett(2023)]{garnett_bayesoptbook_2023}
Roman Garnett.
\newblock \emph{{Bayesian Optimization}}.
\newblock Cambridge University Press, 2023.

\bibitem[Ghassami et~al.(2018)Ghassami, Salehkaleybar, Kiyavash, and Bareinboim]{ghassami2018budgeted}
AmirEmad Ghassami, Saber Salehkaleybar, Negar Kiyavash, and Elias Bareinboim.
\newblock Budgeted experiment design for causal structure learning.
\newblock In \emph{International Conference on Machine Learning}, pages 1724--1733. PMLR, 2018.

\bibitem[Giudice et~al.(2023)Giudice, Kuipers, and Moffa]{giudice2023bayesian}
Enrico Giudice, Jack Kuipers, and Giusi Moffa.
\newblock A bayesian take on gaussian process networks.
\newblock \emph{arXiv preprint arXiv:2306.11380}, 2023.

\bibitem[Glymour et~al.(2019)Glymour, Zhang, and Spirtes]{glymour2019review}
Clark Glymour, Kun Zhang, and Peter Spirtes.
\newblock Review of causal discovery methods based on graphical models.
\newblock \emph{Frontiers in genetics}, 10:\penalty0 524, 2019.

\bibitem[Hauser and B{\"u}hlmann(2014)]{hauser2014two}
Alain Hauser and Peter B{\"u}hlmann.
\newblock Two optimal strategies for active learning of causal models from interventional data.
\newblock \emph{International Journal of Approximate Reasoning}, 55\penalty0 (4):\penalty0 926--939, 2014.

\bibitem[Havercroft and Didelez(2012)]{havercroft2012simulating}
WG~Havercroft and Vanessa Didelez.
\newblock Simulating from marginal structural models with time-dependent confounding.
\newblock \emph{Statistics in medicine}, 31\penalty0 (30):\penalty0 4190--4206, 2012.

\bibitem[Hoyer et~al.(2008)Hoyer, Janzing, Mooij, Peters, and Sch{\"o}lkopf]{hoyer2008nonlinear}
Patrik Hoyer, Dominik Janzing, Joris~M Mooij, Jonas Peters, and Bernhard Sch{\"o}lkopf.
\newblock Nonlinear causal discovery with additive noise models.
\newblock \emph{Advances in neural information processing systems}, 21, 2008.

\bibitem[Jamil and Yang(2013)]{jamil2013literature}
Momin Jamil and Xin-She Yang.
\newblock A literature survey of benchmark functions for global optimisation problems.
\newblock \emph{International Journal of Mathematical Modelling and Numerical Optimisation}, 4\penalty0 (2):\penalty0 150--194, 2013.

\bibitem[Janzing et~al.(2012)Janzing, Mooij, Zhang, Lemeire, Zscheischler, Daniu{\v{s}}is, Steudel, and Sch{\"o}lkopf]{janzing2012information}
Dominik Janzing, Joris Mooij, Kun Zhang, Jan Lemeire, Jakob Zscheischler, Povilas Daniu{\v{s}}is, Bastian Steudel, and Bernhard Sch{\"o}lkopf.
\newblock Information-geometric approach to inferring causal directions.
\newblock \emph{Artificial Intelligence}, 182:\penalty0 1--31, 2012.

\bibitem[Kocaoglu et~al.(2017)Kocaoglu, Dimakis, and Vishwanath]{kocaoglu2017cost}
Murat Kocaoglu, Alex Dimakis, and Sriram Vishwanath.
\newblock Cost-optimal learning of causal graphs.
\newblock In \emph{International Conference on Machine Learning}, pages 1875--1884. PMLR, 2017.

\bibitem[Konobeev et~al.(2023)Konobeev, Etesami, and Kiyavash]{konobeev2023causal}
Mikhail Konobeev, Jalal Etesami, and Negar Kiyavash.
\newblock Causal bandits without graph learning.
\newblock \emph{arXiv preprint arXiv:2301.11401}, 2023.

\bibitem[Lattimore and Szepesv{\'a}ri(2020)]{lattimore2020bandit}
Tor Lattimore and Csaba Szepesv{\'a}ri.
\newblock \emph{Bandit algorithms}.
\newblock Cambridge University Press, 2020.

\bibitem[Lee and Bareinboim(2018)]{lee2018structural}
Sanghack Lee and Elias Bareinboim.
\newblock Structural causal bandits: Where to intervene?
\newblock \emph{Advances in neural information processing systems}, 31, 2018.

\bibitem[Lee and Bareinboim(2019)]{lee2019structural}
Sanghack Lee and Elias Bareinboim.
\newblock Structural causal bandits with non-manipulable variables.
\newblock In \emph{Proceedings of the AAAI Conference on Artificial Intelligence}, volume~33, pages 4164--4172, 2019.

\bibitem[Lorch et~al.(2021)Lorch, Rothfuss, Sch{\"o}lkopf, and Krause]{lorch2021dibs}
Lars Lorch, Jonas Rothfuss, Bernhard Sch{\"o}lkopf, and Andreas Krause.
\newblock Dibs: Differentiable bayesian structure learning.
\newblock \emph{Advances in Neural Information Processing Systems}, 34:\penalty0 24111--24123, 2021.

\bibitem[Lu et~al.(2021)Lu, Meisami, and Tewari]{lu2021causal}
Yangyi Lu, Amirhossein Meisami, and Ambuj Tewari.
\newblock Causal bandits with unknown graph structure.
\newblock \emph{Advances in Neural Information Processing Systems}, 34:\penalty0 24817--24828, 2021.

\bibitem[Malek et~al.(2023)Malek, Aglietti, and Chiappa]{malek2023additive}
Alan Malek, Virginia Aglietti, and Silvia Chiappa.
\newblock Additive causal bandits with unknown graph.
\newblock In \emph{International Conference on Machine Learning}, pages 23574--23589. PMLR, 2023.

\bibitem[Masegosa and Moral(2013)]{masegosa2013interactive}
Andr{\'e}s~R Masegosa and Seraf{\'\i}n Moral.
\newblock An interactive approach for bayesian network learning using domain/expert knowledge.
\newblock \emph{International Journal of Approximate Reasoning}, 54\penalty0 (8):\penalty0 1168--1181, 2013.

\bibitem[Mo{\v{c}}kus(1975)]{movckus1975bayesian}
Jonas Mo{\v{c}}kus.
\newblock On bayesian methods for seeking the extremum.
\newblock In \emph{Optimization Techniques IFIP Technical Conference: Novosibirsk, July 1--7, 1974}, pages 400--404. Springer, 1975.

\bibitem[Murphy(2001)]{murphy2001active}
Kevin~P Murphy.
\newblock Active learning of causal bayes net structure.
\newblock Technical report, technical report, UC Berkeley, 2001.

\bibitem[Ness et~al.(2017)Ness, Sachs, Mallick, and Vitek]{ness2017bayesian}
Robert~Osazuwa Ness, Karen Sachs, Parag Mallick, and Olga Vitek.
\newblock A bayesian active learning experimental design for inferring signaling networks.
\newblock In \emph{Research in Computational Molecular Biology: 21st Annual International Conference, RECOMB 2017, Hong Kong, China, May 3-7, 2017, Proceedings 21}, pages 134--156. Springer, 2017.

\bibitem[Pearl(2009)]{pearl_2009}
Judea Pearl.
\newblock \emph{Causality}.
\newblock Cambridge University Press, 2 edition, 2009.
\newblock \doi{10.1017/CBO9780511803161}.

\bibitem[Shimizu et~al.(2006)Shimizu, Hoyer, Hyv{\"a}rinen, Kerminen, and Jordan]{shimizu2006linear}
Shohei Shimizu, Patrik~O Hoyer, Aapo Hyv{\"a}rinen, Antti Kerminen, and Michael Jordan.
\newblock A linear non-gaussian acyclic model for causal discovery.
\newblock \emph{Journal of Machine Learning Research}, 7\penalty0 (10), 2006.

\bibitem[Spirtes et~al.(2000)Spirtes, Glymour, Scheines, Kauffman, Aimale, and Wimberly]{spirtes2000constructing}
Pater Spirtes, Clark Glymour, Richard Scheines, Stuart Kauffman, Valerio Aimale, and Frank Wimberly.
\newblock Constructing bayesian network models of gene expression networks from microarray data.
\newblock 2000.

\bibitem[Srinivas et~al.(2009)Srinivas, Krause, Kakade, and Seeger]{srinivas2009gaussian}
Niranjan Srinivas, Andreas Krause, Sham~M Kakade, and Matthias Seeger.
\newblock Gaussian process optimization in the bandit setting: No regret and experimental design.
\newblock \emph{arXiv preprint arXiv:0912.3995}, 2009.

\bibitem[Surjanovic and Bingham(2013)]{surjanovic2013drop}
S~Surjanovic and D~Bingham.
\newblock Drop-wave function.
\newblock 2013.

\bibitem[Sussex et~al.(2022)Sussex, Makarova, and Krause]{sussex2022model}
Scott Sussex, Anastasiia Makarova, and Andreas Krause.
\newblock Model-based causal bayesian optimization.
\newblock \emph{arXiv preprint arXiv:2211.10257}, 2022.

\bibitem[Tigas et~al.(2022)Tigas, Annadani, Jesson, Sch{\"o}lkopf, Gal, and Bauer]{tigas2022interventions}
Panagiotis Tigas, Yashas Annadani, Andrew Jesson, Bernhard Sch{\"o}lkopf, Yarin Gal, and Stefan Bauer.
\newblock Interventions, where and how? experimental design for causal models at scale.
\newblock \emph{Advances in Neural Information Processing Systems}, 35:\penalty0 24130--24143, 2022.

\bibitem[Tigas et~al.(2023)Tigas, Annadani, Ivanova, Jesson, Gal, Foster, and Bauer]{tigas2023differentiable}
Panagiotis Tigas, Yashas Annadani, Desi~R Ivanova, Andrew Jesson, Yarin Gal, Adam Foster, and Stefan Bauer.
\newblock Differentiable multi-target causal bayesian experimental design.
\newblock In \emph{International Conference on Machine Learning}, pages 34263--34279. PMLR, 2023.

\bibitem[Tong and Koller(2001)]{tong2001active}
Simon Tong and Daphne Koller.
\newblock Active learning for structure in bayesian networks.
\newblock In \emph{International joint conference on artificial intelligence}, volume~17, pages 863--869. Citeseer, 2001.

\bibitem[Toth et~al.(2022)Toth, Lorch, Knoll, Krause, Pernkopf, Peharz, and Von~K{\"u}gelgen]{toth2022active}
Christian Toth, Lars Lorch, Christian Knoll, Andreas Krause, Franz Pernkopf, Robert Peharz, and Julius Von~K{\"u}gelgen.
\newblock Active bayesian causal inference.
\newblock \emph{Advances in Neural Information Processing Systems}, 35:\penalty0 16261--16275, 2022.

\bibitem[Verma and Pearl(2022)]{verma2022equivalence}
Thomas~S Verma and Judea Pearl.
\newblock Equivalence and synthesis of causal models.
\newblock In \emph{Probabilistic and causal inference: The works of Judea Pearl}, pages 221--236. 2022.

\bibitem[von K{\"u}gelgen et~al.(2019)von K{\"u}gelgen, Rubenstein, Sch{\"o}lkopf, and Weller]{von2019optimal}
Julius von K{\"u}gelgen, Paul~K Rubenstein, Bernhard Sch{\"o}lkopf, and Adrian Weller.
\newblock Optimal experimental design via bayesian optimization: active causal structure learning for gaussian process networks.
\newblock \emph{arXiv preprint arXiv:1910.03962}, 2019.

\bibitem[Wang and Jegelka(2017)]{wang2017max}
Zi~Wang and Stefanie Jegelka.
\newblock Max-value entropy search for efficient bayesian optimization.
\newblock In \emph{International Conference on Machine Learning}, pages 3627--3635. PMLR, 2017.

\bibitem[Weiss(2012)]{weiss2012elementary}
Neil~A Weiss.
\newblock \emph{Elementary statistics}.
\newblock New York, 2012.

\bibitem[Williams and Rasmussen(1995)]{williams1995gaussian}
Christopher Williams and Carl Rasmussen.
\newblock Gaussian processes for regression.
\newblock \emph{Advances in neural information processing systems}, 8, 1995.

\bibitem[Yang et~al.(2018)Yang, Katcoff, and Uhler]{yang2018characterizing}
Karren Yang, Abigail Katcoff, and Caroline Uhler.
\newblock Characterizing and learning equivalence classes of causal dags under interventions.
\newblock In \emph{International Conference on Machine Learning}, pages 5541--5550. PMLR, 2018.

\end{thebibliography}

\newpage

\appendix
\onecolumn
\section{APPENDIX}

\subsection{Nomenclature}
Using standard notation, we use Capital letters to denote random variables and lowercase letters to denote the realization of said random variables. We use bold letters to denote sets of certain nodes. The support of a variable is given by curly letters. We use the subscript $t$ to index data observed thus far, and the subscript $i$ is used to index a particular node in a vector, the superscript $g$ is used to refer to the input space 

\label{sec:nomenclature}

\begin{table}[htbp]
\centering
\renewcommand{\arraystretch}{1.2} 
\begin{tabular}{ccl}
\textbf{Symbol} &  & \textbf{Description} \\
 & & \\
$V_j$ & & $j^{th}$ observed variable\\
$Y$ & & Target variable we seek to optimize corresponds to $V_m$ \\
$\bm{V}$ & & Set of all observed variables \\
$\bm{X}$ & & Set of intervenable variables \\
$\bm{C}$ & & Set of non intervenable variables \\
$A_i$ & & Action performed on node i \\
${g^*}$ & & True latent causal graph \\
$\bm{A}$ & & Action vector composed of $\{A_i\}_{i=0}^m$ \\
$\bm{Z}_i^{g^*}$ & & The parents of node $i$ in graph ${g^*}$ \\
$f_i^{g^*}(\bm{z}_i^{g^*},\bm{a}_i^{g^*})$ & & functions relating a node $i$ with it's parents and actions \\
$\bm{f}^{g^*}(\bm{a})$  & & The overall function with input action composed of functions $\{f_i^{g^*}\}_{i=0}^{m}$ related by graph ${g^*}$ \\
$\bm{F}_{g^*}$ & & the set of respective unknown functions associated with $g^*$, i.e. $\{f_i^{g^*}\}_{i=0}^m$\\
$\boldsymbol{\Omega}$ & & a set of independent noises with zero mean and known distribution, i.e. $\{\Omega_i\}_{i=0}^m$\\
$pa_{g^*}(i) $ & & indices of parent nodes of any node, defined for the \textsc{dag} ${g^*}$\\
$\{y_t,\bm{v}_t,\bm{a}_t\}$ & & Observation of reward variable $y_t$ and intermediate variables $v_t$ for the corresponding action $\bm{a}_t$ \\
$\mathcal{D}_t$ & & Observations for actions until time $t$, is the set $\{y_j,\bm{v}_j,\bm{a}_j\}_{j=0}^{t}$\\
$G_t$ & & Posterior of the distribution over graphs at time $t$ \\
$g$ & & Random DAG samples from $G_t$ \\
$k_i^g(\cdot,\cdot)$ & & Kernel defined on input space implied by graph $g$ for node $i$, gives covariance between two points \\
$\bm{k}_{i,t}^g(\cdot)$ & & A vector of covariances of the current input to previous inputs $[k_i^g((z_{i,t}^g,a_{i,t}^g),\cdot)]_{i=0}^t$ \\
$\bm{K}_{i}^g$ & & Covariance matrix based on previous $\mathcal{D}_t$\\
$\mu_{i,t}^g(\cdot)$ & & Mean function based on data $\mathcal{D}_t$ and kernel $k_i^g(\cdot,\cdot)$\\
$\sigma_{i,t}^g(\cdot)$ & & Variance function based on data $\mathcal{D}_t$ and kernel $k_i^g(\cdot,\cdot)$ \\
$GP(\mu_{i,t}^g, \sigma_{i,t}^g)$ & & Gaussian Process $\Tilde{f}_i^g(\cdot) \sim \mathcal{N}(\mu_{i,t}^g(\cdot),\sigma_{i,t}^g(\cdot))$\\
$\mathcal{H}_{k_i^g}$ & & Hilbert Spaces of functions implied by kernel $k_i^g$ \\
$\Tilde{f}_i^g$ & & A function sampled from Gaussian Processes GP \\ 
$\omega_i$ & & Observational noise of node $i$ \\
$\mathcal{M}_t$ & & Plausible models at time $t$ based on confidence bounds

\end{tabular}
\label{tab:TableOfNotation}
\end{table}


\subsection{Simulation Details}
\label{sec:SimulationDetails}

\paragraph{Dropwave} \cite{surjanovic2013drop,astudillo2021bayesian,sussex2022model} \\
For our setting we consider $A_0\in[-5.12,5.12]$ and $A_1\in[-5.12,5.12]$, we set $\beta=0.5$ and $\epsilon_i=0.1 \forall i\in[m]$
\begin{figure}[ht]
    \centering
    \includegraphics[scale=0.35]{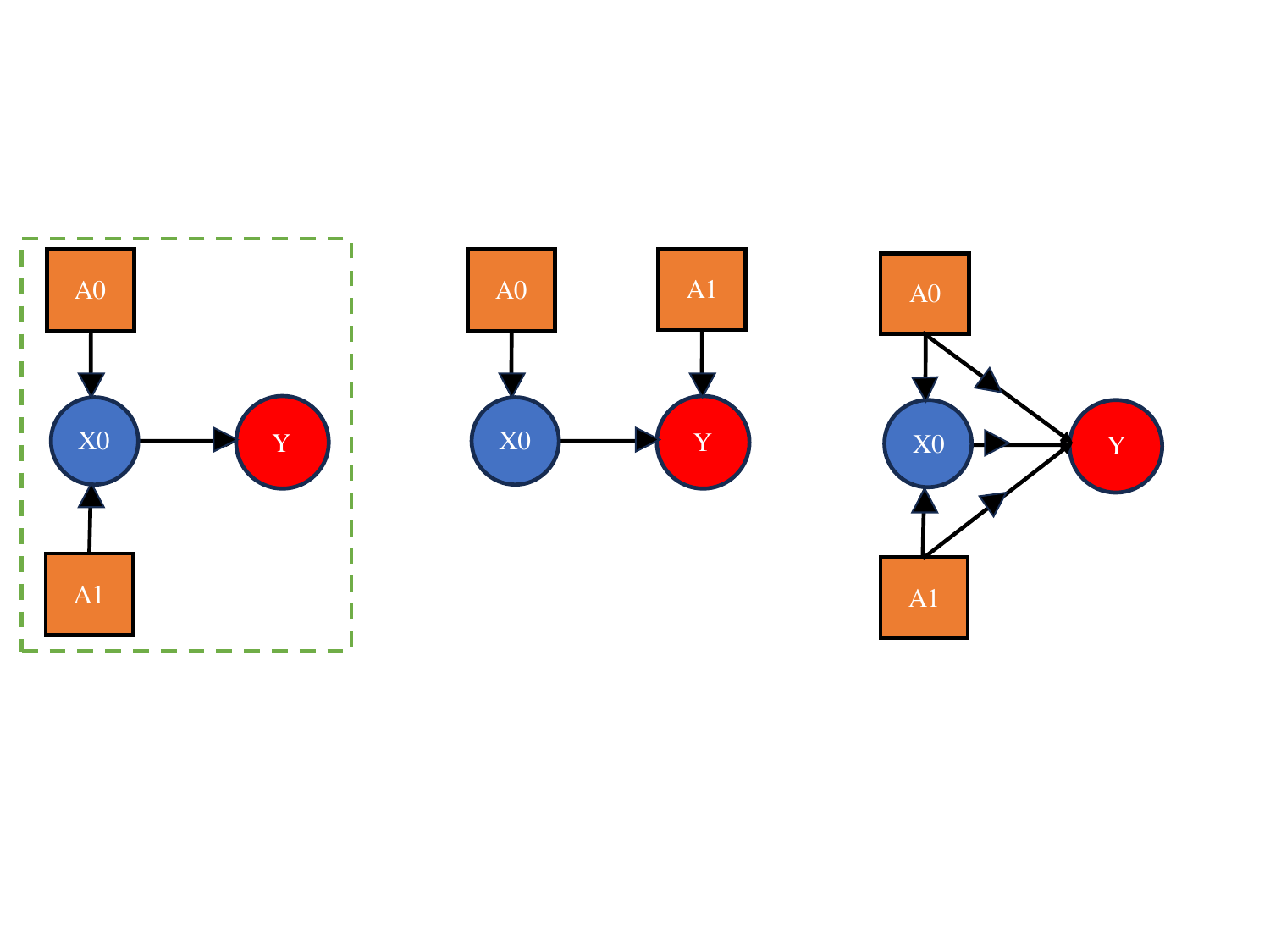}
    \caption{Dropwave: True DAG structure, and Incorrect DAG structures used in Experiment}
    \label{fig:Dropwave, correct and wrong}
\end{figure}

\begin{equation}
    \begin{aligned}
        x_0 = f_0(a_0,a_1) &= \sqrt{a_0^2 + a_1^2} + \epsilon_0\\
        y = f_y(x_0) &= \frac{1 + \cos(12x_0)}{2 + 0.5x_0^2}+\epsilon_y
    \end{aligned}
\end{equation}
\paragraph{Rosenbrock}
\cite{jamil2013literature,astudillo2021bayesian,sussex2022model}\\
For our setting we use $a_i \in [-2,2]$ for $i \in [m]$, we use $\beta=0.5$ and $\epsilon_i=0.1\forall i \in [m]$
\begin{figure}[h]
    \centering
    \includegraphics[scale=0.35]{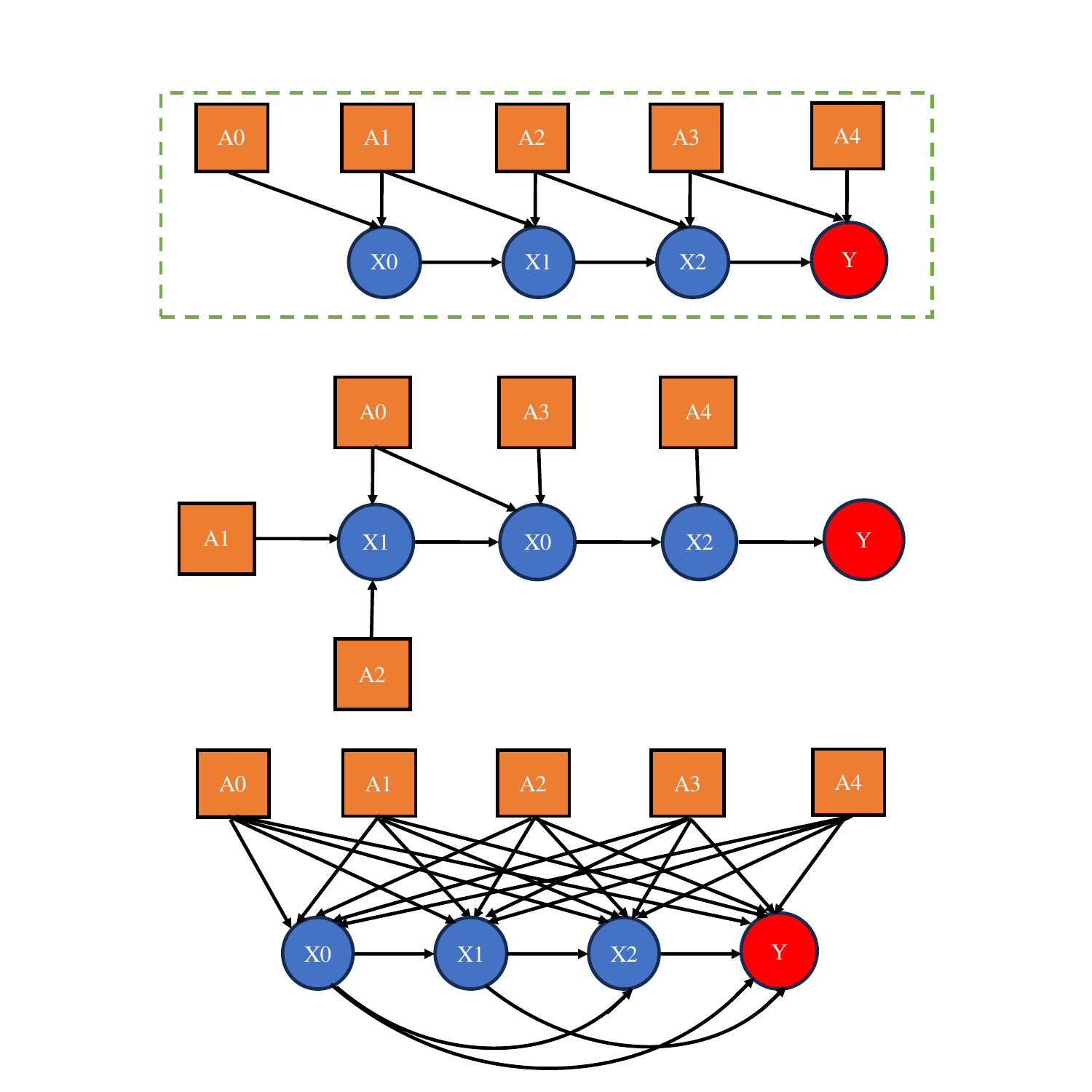}
    \caption{Rosenbrock: True DAG structure, and Incorrect DAG structures used in Experiment}
    \label{fig:Rosenbrock, correct and wrong}
\end{figure}
\begin{equation}
    \begin{aligned}
        f_0(a_0,a_1) &= -100{a_1-a_0^2}2 - (1-a_0)^2 + \epsilon_0 \\
        f_k(a_k,a_{k+1},x_{k-1}) &= -100{a_{k+1}-a_k^2}2 - (1-a_k)^2 + x_{k-1} + \epsilon_k i=1,\dots,m
    \end{aligned}
\end{equation}
\paragraph{Alpine3}\citep{jamil2013literature,sussex2022model,astudillo2021bayesian}
For our experiments we consider $a_i\in[0,10]$, for $i\in m$, we consider $\beta=0.5$ and $\eta=0.1$
\begin{figure}[h]
    \centering
    \includegraphics[scale=0.35]{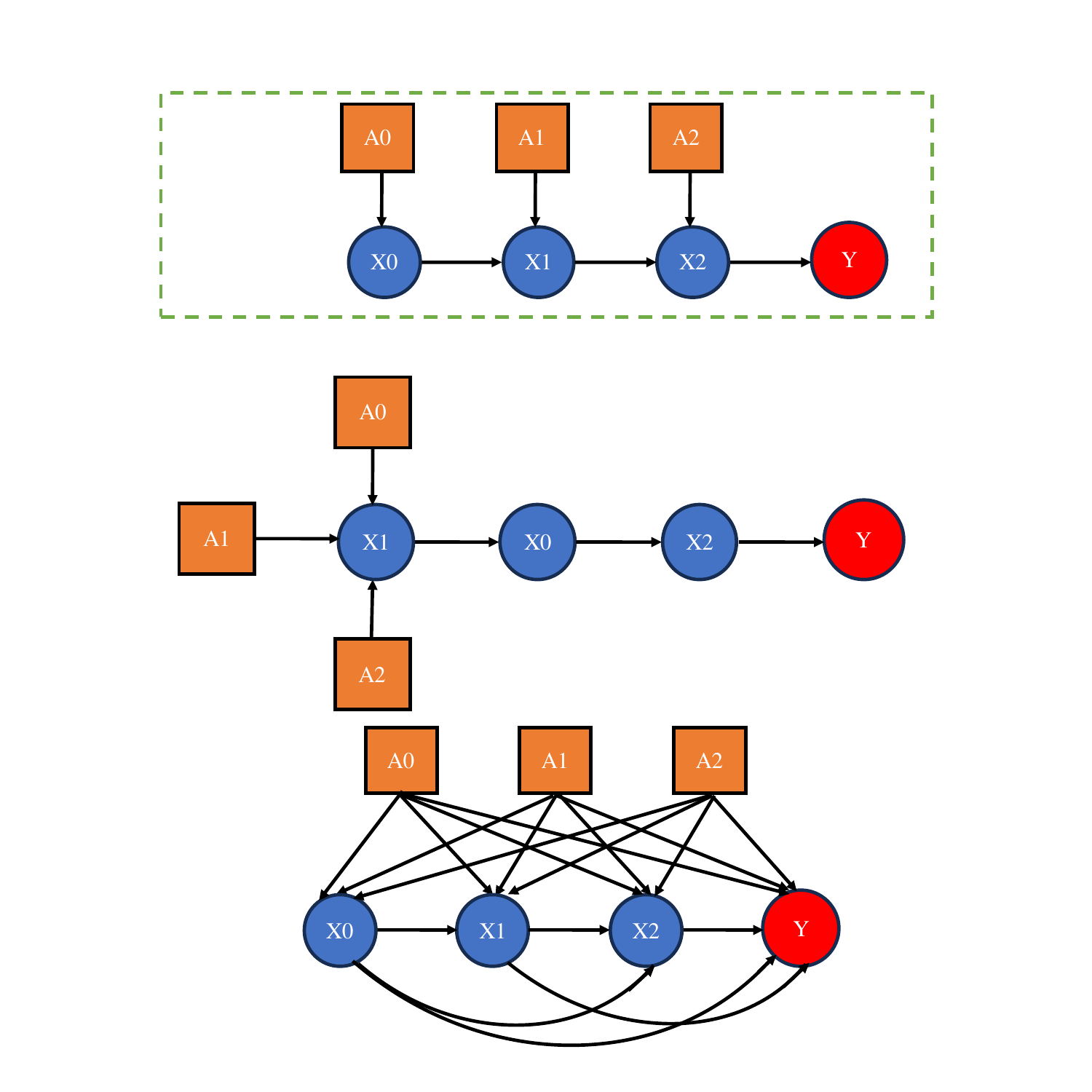}
    \caption{Alpine3: True DAG structure, and Incorrect DAG structures used in Experiment}
    \label{fig:Alpine3, correct and wrong}
\end{figure}
\begin{equation}
    \begin{aligned}
        f_0(x_0) &= -\sqrt{x_0}\sin(x_0) + \epsilon_0\\
        f_i(a_i,x_{i-1}) &= \sqrt{a_i}\sin(a_i)x_{i-1} + \epsilon_i, i=1,\dots,m
    \end{aligned}
\end{equation}

\paragraph{ToyGraph}\cite{aglietti2020causal}
\begin{figure}[h]
    \centering
    \includegraphics[scale=0.35]{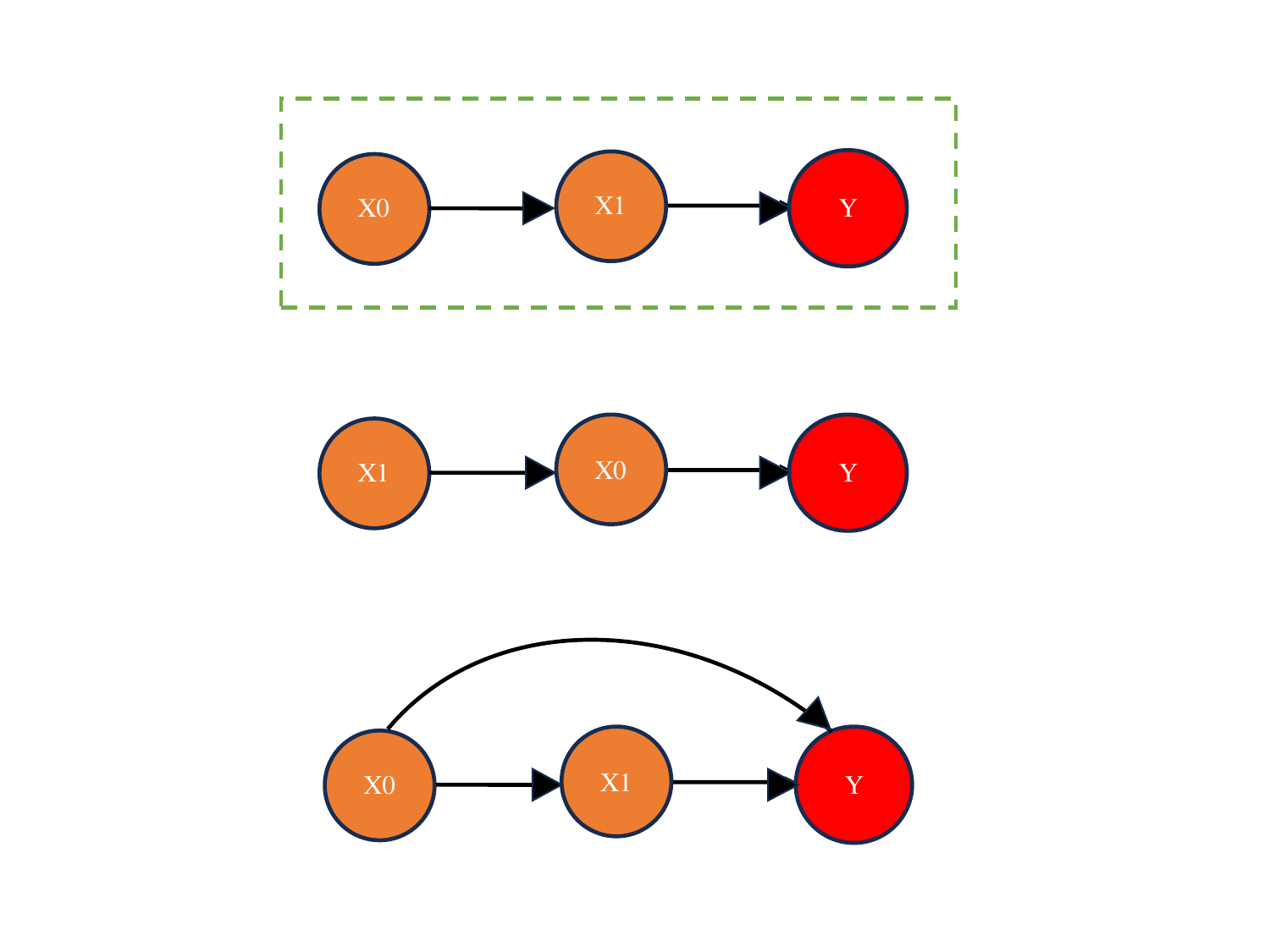}
    \caption{Toy Graph: True DAG structure, and Incorrect DAG structures used in Experiment}
    \label{fig:Toy Graph, correct and wrong}
\end{figure}
\begin{equation}
    \begin{aligned}
        X &= \epsilon_x \\ 
        Z &= \exp(-X) + \epsilon_z \\
           Y &= \cos(Z) - \exp(-Z/20) + \epsilon_y 
    \end{aligned}
\end{equation}
\paragraph{Epidemiology}\cite{branchini2023causal}\cite{havercroft2012simulating}\\
In our settings, we consider the following input ranges for interventions
$T \in [0,4]$ and $R \in [0,4]$, we use $\beta=1$ and noise levels are specified according to the \textsc{scm}.

\begin{equation}
    \begin{aligned}
        B &= \mathcal{U}[-1,1] \\ 
        T &= \mathcal{U}[4,8] \\ 
        L &= \text{expit}(0.5T + U) \\
        R &= 4 + L T \\
        Y &= 0.5 + \cos(4 T) + \sin(-L + 2 R) + B + \epsilon \text{ with} \epsilon\sim\mathcal{N}(0,1)
    \end{aligned}
\end{equation}

\subsection{ Causal Discovery and Causal Bayesian Optimization}
\label{sec: ACD - CBO}

Causal discovery from observational data \citep{verma2022equivalence,andersson1997characterization,spirtes2000constructing,chickering2002optimal,friedman2003being,shimizu2006linear} can recover causal graphs up to Markov Equivalence Classes (\textsc{mec}). \cite{friedman2003being,janzing2012information} go beyond \textsc{mec} from purely observational data based on information asymmetry. \cite{tong2001active,murphy2001active,eaton2007exact,hauser2014two,wang2017max,ness2017bayesian,yang2018characterizing,ghassami2018budgeted,agrawal2019abcd,faria2022differentiable} study the problem of learning graphs from observational and interventional data. 

Active causal discovery aims to learn the \textsc{scm} efficiently. For example, \cite{von2019optimal} studied causal structure learning actively using Bayesian Optimal Experimental Design (\textsc{boed}). The acquisition function used in their model seeks to select the intervention that is maximally informative about the underlying causal structure with respect to the current model. 

Bayesian optimisation aims to learn the optimal point of unknown functions.
Knowing the causal structure helps us reduce the causal intrinsic dimension of the optimisation problem for hard intervention. In the case of soft interventions, causal knowledge is useful for utilising the information from intermediate nodes and converting a high dimension problem into $n$ smaller dimensional optimisation problems (where $n$ is the number of intermediate nodes).

However, learning the entire \textsc{scm} such as in active causal discovery (i.e., all the causal edges and mechanisms of all nodes in their entire domain) is not necessary for causal Bayesian optimisation.
This is understood in two separate cases:

\paragraph{Hard Intervention}
A hard Intervention mutates the graph, making the intervened node independent of all its parents and ancestors. \cite{lee2018structural} demonstrated that the optimal intervention lies within the parents when there are no unobserved confounders. In such a case learning the causal relation between the ancestors of the parents does not help the underlying goal of causal Bayesian optimisation.
Consider the example of ToyGraph, in the true data-generating mechanism $X_1$ is the parent of $Y$, and on performing $\text{do}(X_1=x_1)$ the value of $Y$ is not affected by the value of $X_0$, hence for optimization knowing the causal direction or mechanism relating $X_0$ and $X_1$ is not required.

\paragraph{Soft Intervention}
A soft intervention does not mutate the graph, hence learning the causal relations of the ancestral nodes is still relevant to the downstream optimisation problem, however learning the entire causal structure might still be wasteful. If we have determined (specified by expert knowledge or during a certain step of our active causal discovery process that a certain node is not an ancestor of the target node, then knowing the ancestors or descendants of the node does not contribute to causal Bayesian optimisation. 
Causal structure in the case of soft intervention utilises values of intermediate nodes to constrain the optimisation problem. Causal structure is only useful when the decomposed problem is simpler than the original problem. For example consider function $f(x_1,x_2) = g(h(x_1),x_2)$, knowing the intermediate value $h(x_1)$ is only useful if the composed function $f$ is more difficult to optimise (because of non-linearity) than the individual functions $g$ and $h$. 
We observe in our experiments with the Rosenbrock graph in Section \ref{sec: experiments} that there is no significant advantage of causal structure when the intermediate functions are purely linear.

The goals of active causal discovery and Bayesian optimisation are misaligned. While Bayesian optimisation tries to balance exploration and exploitation to minimise cumulative regret, the active causal discovery acquisition function might choose an intervention that has a low reward and does not help future steps of causal Bayesian optimisation but helps discover the true underlying Causal Graph.  
Therefore it is sub-optimal to firstly perform active causal discovery and then followed by causal Bayesian optimisation as separate steps. Since the active causal discovery step might spend additional samples with low reward learning useless parts of the Causal graph. 

Our algorithm naturally unifies these two steps by making causal discovery a sub-task of causal Bayesian optimisation.
If multiple causal graphs exist within our hypothesis space that explains the data collected up to time step $t$, we only perform an intervention aimed at disambiguation between these graphs if it potentially leads to better rewards than the optimal value observed thus far. Our acquisition function \eqref{alg:gacbohard} has three maximisation, for a given graph $g \in G_t$ there are several plausible functions for each node $\Tilde{f}_{i,t}^g$ and all possible combination of node functions define the function space for the graph $g$. We use the optimistic reparameterisation trick to find the combination of functions and actions $a_g$ which maximises the target node. We do this for all plausible graphs and compare the best possible value for each graph $g$. We select the plausible graph $g$ with the maximum possible value for target node and the corresponding action $a_g$ which maximises it.
Consider a hypothetical scenario with two different graphs $g_1,g_2$ which disagree on the value of node $i$ for intervention $a$ but the action which maximises the value of target node $y_{g_1}, y_{g_2}$ in $g_1$ and $g_2$ is same $a^*$, and the target node values also agree i.e., $y_{g_1}^* = y_{g_2}^*$ for action $a^*$, even though performing $a$ would help identify the true graph our acquisition function is designed to choose $a^*$. Because $y_{g_1} \leq y_{g_1}^*$ or $y_{g_2} \leq y_{g_2}^*$ for any action $a\neq a^*$.

\subsection{Superexponential Scaling of DAGs and Scalability}
\label{sec: Scalability}
The problem setting we addressed in this paper is challenging due to the super-exponential growth of the number of DAGs with the increase in the number of nodes. We only focus on small graphs where all the graphs can be enumerated to study the problem of \textsc{CBO} with unknown graphs in isolation. Our approach can be further improved to be more scalable, by MCMC-based sampling in the space of graphs \cite{giudice2023bayesian}, or Differential approaches like DiBS \cite{lorch2021dibs} in latent spaces or topological ordering of nodes. We leave it as a future work.

Our method in its current state computes the \textsc{gp} score of all possible graph components (all combinations of parents and actions for all observed nodes) and samples graphs based on the GP score and individually optimises and compares all sampled graphs. For larger graphs, the problem becomes intractable as the number of components for which the \textsc{gp} score needs to be calculated increases exponentially.
In the initial rounds, the number of graphs that need to be optimised and the number of comparisons that need to be made also increases superexponentially. Causal Bayesian optimisation using the MCBO approach also takes longer for larger graphs. 

\subsection{Discussion on Theoretical Analysis}
\label{sec: Discussion on Theoretical Analysis}
Our approach suggests attaining a similar regret bound to MCBO but with an added constant term. However, our method holds potential for a superior regret bound by simultaneously exploring causal structure and exploiting rewards from the outset. Empirical results indicate that our algorithm, GACBO, exhibits a significantly faster increase in average rewards after initial rounds, underscoring its potential for improved regret.
The lack of guarantees for the convergence of the posterior to the true graph in finite samples is a major obstacle. 
A potential theoretical proofing can be achieved by decomposing our regret into two parts: 
\begin{itemize}
    \item Constant term: For first $n$ samples before learning the true graph, we obtain the constant regret. This is due to the boundness assumption of function (See section 2.3 “Regularity Assumption”) $||f_i^{g^\ast}|| \leq \mathcal{B}_i$. No matter what actions are selected, the upper bound of instant regret can be bounded by $2\mathcal{B}_i$. 
    \item MCBO regret: the second term is the same as the MCBO regret term since after $n$ samples we’ve discovered the true graph. 
\end{itemize}

\paragraph{Effect of Graph Knowledge on Optimisation} Theorem 1 of \cite{sussex2022model} bounds the regret with high probability when the graph is known but functions are unknown in the case of
soft intervention as $R_T \leq \mathcal{O}(L_f^NL_\sigma^N\beta_T^NK^Nm\sqrt{T\gamma_T})$ where $\gamma_T=\max_i \gamma_{i,T}$, and $N$ denotes the maximum distance from a root node to $V_m$, $K = \max_i|pa_g^*(i)|$  as compared to Standard Bayesian Optimisation that makes no use of graph structure resulting in cumulative regret \textit{exponential} in $m$. Assuming the use of the Squared Exponential Kernel for modeling all functions, $\gamma_T=\mathcal{O}((K+q)(logT)^{K+q+1})$ scales exponentially with respect to $K$ and $q$ the length of each action vector. This results in an expression that scales exponentially in $K,N$.
The theorem demonstrates a potentially exponential improvement in the scaling of cumulative regret for possible actions $m\geq K+N$.

For environments allowing hard interventions, the optimisation problem can be reduced to the Causal Intrinisic dimension \citep{aglietti2020causal} if interventions on parents is allowed, \cite{lee2018structural} shows results that the optimal intervention is always found among parents.For environments which do not allow direct interventions on parents
the problem can be studied as a soft intervention for the mutated graph, by treating the intervened nodes as action nodes and propagating uncertainty through the remaining nodes. For certain graphs the depth $N$ of the resulting graph can be reduced significantly, consider by figure 1 of \cite{aglietti2020causal} with a slight modification where, intervention on the parent nodes $\{X_{100},Z_{100}\}$ is not possible but we are allowed to intervene on $\{X_{99},Z_{99}\}$, this allows us to reduce $N=2$ from $N=100$, resulting in an exponential improvement in performance. 

\paragraph{Convergence to the True Graph} As the posterior mass on the graph distribution converges to the dirac delta distribution on the true graph $p(g) \xrightarrow{} \delta_{g=g^*}$the cumulative regret converges to the cumulative regret accrued when the graph is known. For hard interventions the posterior convergence to the true graph is guaranteed under a few assumptions. 
For soft intervention models \textsc{dag}s belonging to Markov Equivalence Classes are further distinguished under the assumptions underpinning the \textsc{gpn} models, considering the functions $f_i$, are not generally invertible \citep{giudice2023bayesian}, the \textsc{gpn} usually suggest higher scores to models admitting the true \textsc{sem} structure as confirmed in our numerical experiments. For cases where functions are invertible \citep{hoyer2008nonlinear} guarantees identifiability by leveraging the asymmetry of residual noise distributions.

While asymptotic convergence to the true graph structure is guaranteed, there are no known results for finite samples. 
However, in our numerical experiments, we observe that the graph converges to the essential graph in a small number of samples and potentially observes exponentially less regret as compared to not knowing the graph henceforth.
Several studies have considered the problem of learning the causal structure optimally, \cite{murphy2001active, tong2001active,masegosa2013interactive,hauser2014two,kocaoglu2017cost}.
Future work could look at more efficient techniques to learn the structure with finite time guarantees to place an upper bound on the cumulative regret for the case when the graph is unknown apriori. 

\subsection{Experiment Details} All our experiments were performed on Google Colab without a GPU or TPU enabled, we used random seeds $47,42,73,66,13$ for 5 repeat for all given algorithms and given environments. 

\subsection{Limitations and Future Work} In the current work we focus on the problem of causal Bayesian optimisation with unknown graph, however we make several assumptions which may be violated in practise. We assume no unobserved confounders, this assumption is critical to our causal discovery algorithm and our model based approach for causal Bayesian optimisation is also not resilient to unobserved confounders. We assume additive noise and known noise distribution for each node, this is a strong assumption in practise and needs to be relaxed in future work. Our regularity assumptions might also restrict the application of our method to problems where the relation between a node and it's parents is not highly nonlinear. Our current method does not scale well to larger graphs, however this can be addressed in future work as described in section \ref{sec: Scalability}. We defer providing theoretical guarantees for our method to future work as discussed in \ref{sec: Discussion on Theoretical Analysis}.

\end{document}